\documentclass{article}

\usepackage{arxiv}

\usepackage[utf8]{inputenc} 
\usepackage[T1]{fontenc}    
\usepackage{hyperref}       
\usepackage{url}            
\usepackage{booktabs}       
\usepackage{amsfonts}       
\usepackage{nicefrac}       
\usepackage{microtype}      
\usepackage{lipsum}		
\usepackage{graphicx}
\usepackage{natbib}
\usepackage{doi}
\usepackage{amsmath}
\usepackage{algorithm}
\usepackage{algpseudocode}
\usepackage{xcolor}
\usepackage{subcaption}
\usepackage{etoc}
\newcommand{\ours}{StateCraft}

\title{StateCraft: Augmenting Video World Models with Abstract-State Reasoning}

\author{%
  \parbox{\textwidth}{\centering
    Lorenzo Venturoli\textsuperscript{1,2}\thanks{Corresponding author: \texttt{lorenzo.venturoli@insait.ai}},\;
    Nedko Savov\textsuperscript{1},\;
    Xi Wang\textsuperscript{3,4,5},\;
    Marc Pollefeys\textsuperscript{1,2,6},\;
    Luc Van Gool\textsuperscript{1}
    \\[1.0em]
    {\normalfont
    \textsuperscript{1}INSAIT, Sofia University "St. Kliment Ohridski" \quad
    \textsuperscript{2}ETH Zurich \quad
    \textsuperscript{3}University of Ulm \quad
    \textsuperscript{4}TUM \quad \\
    \textsuperscript{5}MCML - Munich Center for Machine Learning \quad
    \textsuperscript{6}Microsoft}
  }
}

\date{}

\renewcommand{\undertitle}{}

\hypersetup{
pdftitle={StateCraft: Augmenting Video World Models with Abstract-State Reasoning},
pdfauthor={Lorenzo Venturoli, Nedko Savov, Xi Wang, Marc Pollefeys},
pdfkeywords={World Models, Memory, VLM, Graphs},
}

\begin{document}
\maketitle

\vspace{-5mm}
\begin{figure}[h]
\centering
\includegraphics[width=\linewidth]{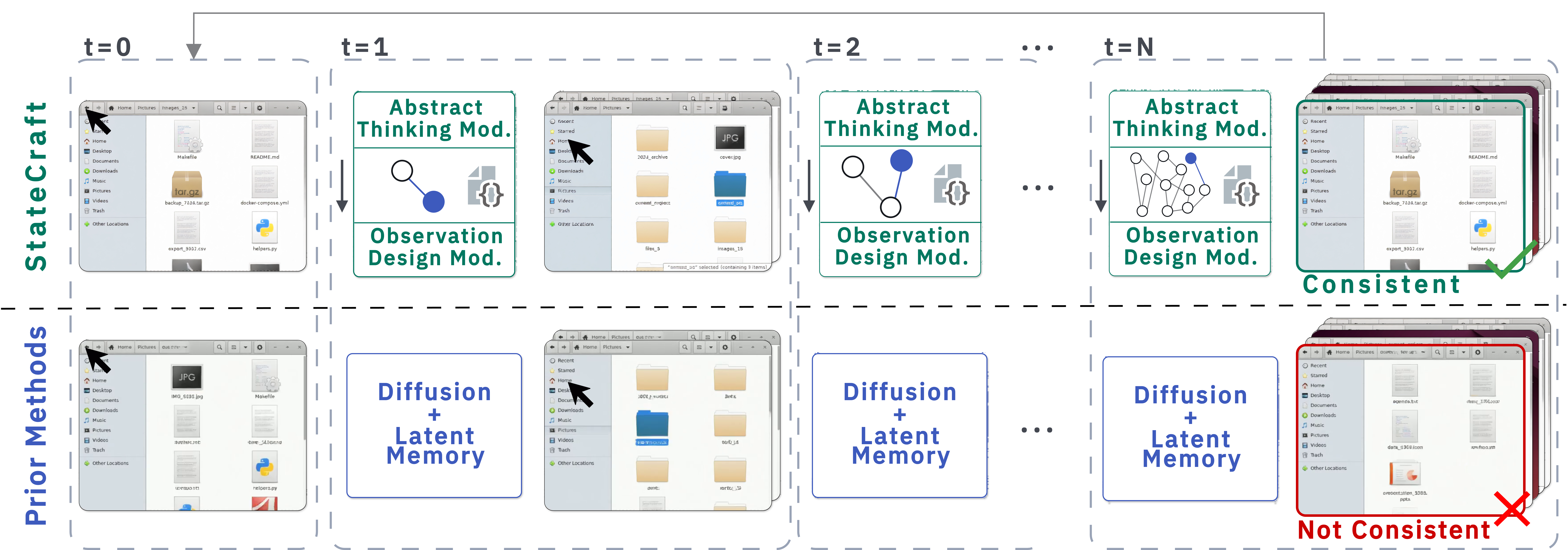}
\caption{\textbf{Abstract and visual state decoupling for consistent video world models.}
\ours{} (top) couples a VLM that reasons about action effects and updates an abstract state with a video world model that generates the corresponding observations. This state preserves information beyond the current view, enabling consistent folder revisits, while prior methods relying on latent memory (bottom) may generate plausible but inconsistent content.}

\label{fig:teaser}
\end{figure}

\begin{abstract}
Video world models aim to simulate complex environments with realistic appearance and dynamics.
Maintaining consistency over long interactions requires tracking past actions and understanding how those actions affect the environment. Video world models can only generate partial observations of the entire world at a time.
Remembering past observations alone may not be sufficient in every scenario, as they may no longer reflect the world after a sequence of actions.
We therefore introduce a paradigm that decouples a persistent, updatable abstract state from the visual state.
The abstract state keeps track of the entities' changes as the world evolves, and is used to condition the visual state, which represents the current view.
We propose \emph{\ours{}}, a novel architecture that pairs a language model with a video diffusion model to evolve the dynamic world.
We evaluate our approach on generative interfaces, which pose a substantial challenge because a single action may change parts of the state that are not currently visible.
To support training and evaluation, we develop a data collection pipeline that produces informative, densely annotated trajectories at scale without human or agent supervision, together with a benchmark measuring generation quality, memory, and controllability. On this benchmark, our approach improves visual memory metrics by 0.30 (SSIM) and PSNR by up to 10 dB over the state-of-the-art baselines.
We demonstrate that \ours{} maintains high self-consistency and strong generation quality over long trajectories. 
Because the abstract state is explicit, it can be edited directly, and generation follows edits to states never observed in the data.
Finally, a proof-of-concept 3D experiment, in which actions change the illumination of an unobserved room, shows that the paradigm can also be applied to spatial environments.
Dataset, code, and model weights will be made public: \url{https://insait-institute.github.io/StateCraft/}.

\end{abstract}

\section{Introduction}
\label{ch:intro}

World models learn the dynamics of a system to predict state changes in response to actions. Recent advances have enabled the simulation of realistic 3D worlds \citep{bruce_genie_2024, nvidia2026cosmos3omnimodalworld}, robotics \citep{yang_learning_2024}, driving scenes \citep{russell2025gaia2controllablemultiviewgenerative, gao2024vistageneralizabledrivingworld}, games \citep{valevski_diffusion_2025, qian2026matrixgame35enhancingrealtime}, and, recently, graphical user interfaces (GUI) \citep{rivard_neuralos_2025, zhuge2026neuralcomputers}. As world models generate only part of the world at a time, actions may lead to changes that are not directly visible, and consistent simulation requires understanding their effects on the world state and updating it accordingly. Current video world models represent this state visually, through frame histories \citep{zhang2025framecontextpackingdrift, cai2025mixturecontextslongvideo}, caches of encoded frames \citep{xiao2025worldmemlongtermconsistentworld, chen2026out}, reconstructed 3D point clouds \citep{shen2026lyra}, or recurrent states trained to reconstruct frames \citep{savov2025statespacediffuserbringinglongcontext}. Most of these memories store static content and cannot be updated by out-of-view dynamics. LiveWorld \citep{duan2026liveworld} supports such dynamics only for the motion of recognized entities within a fixed 3D scaffolding.

The need for a more abstract and general state representation is especially notable in the field of generative software, where each frame is conditioned on a keyboard or mouse event to simulate a GUI. For example, deleting a directory's contents empties the folder, yet the model should track that 'Trash' now contains those files and display them when visited. Existing memory approaches mainly target 3D worlds, and those requiring camera poses or explicit geometry do not directly apply to GUI simulation. Moreover, observations alone do not fully describe the world state behind the scenes, making the simulation visually plausible but not functional in practice \citep{rivard_neuralos_2025}.

To model a more complete picture of the world state, we factorize it into an abstract state and a visual state. The abstract state is not tied to a specific representation. It is any structured record of the environment's entities and their properties, including those out of view, that actions update directly. We develop \ours{}, which instantiates this state as a graph and complements a video world model with a VLM that updates and reasons over it. We demonstrate it on generative software, training the system to simulate a file explorer that preserves filesystem state and dynamics and supports editing them. We further propose trainable triggers on top of the video world model that decide when the VLM updates the abstract state, enabling programmable, event-driven interaction between the two. Unlike history-retrieval methods, our memory is dynamic rather than static. Unlike WorldDirector \citep{wang2026worlddirector}, trajectories are not planned in advance. Our VLM predicts state transitions online in response to incoming actions.

For training and evaluation, we develop a data collection framework for generative software. Existing pipelines combine a few goal-driven trajectories with many random, uninformative ones. Ours selects actions from interface elements and records aligned frames, actions, and structured state changes, producing informative, densely annotated trajectories at scale. We also release, to our knowledge, the first benchmark for video world models of graphical interfaces, evaluating generation quality, memory across controlled revisit horizons, and controllability under explicit state edits. On this benchmark, \ours{} keeps revisit self-consistency near 0.9 SSIM over up to 512 actions, while baselines remain at 0.5--0.6, and improves PSNR by up to 10 dB over the baselines. Direct edits to the abstract state are followed in 81\% of cases, including hidden changes out of view, although inserting new elements on empty background remains difficult. A proof-of-concept 3D experiment shows that abstract state also guides generation beyond software interfaces. These results support modeling abstract state alongside visual dynamics as a basis for consistent, controllable simulation.

Our contributions are the following:
\begin{itemize}
\item A novel paradigm that factorizes the world into a global, abstract state that tracks and reasons about structured changes of the environment, and a visual state that simulates what is in view at every instant, enabling long, consistent, and controllable interface simulation and extending to state-guided generation in 3D.
\item \ours{}, a novel architecture that couples a language model and a video world model throughout ongoing interaction, jointly predicting abstract state transitions and visual observations as user actions arrive, and enables event-driven trajectory control. 
\item A grounded data collection pipeline that records aligned frames, actions, and structured state changes, producing substantially more informative interactions. 
\item A novel benchmark for user-interface world models that isolates memory across controlled revisit intervals and evaluates state prediction, content reconstruction, and generation under explicit state edits. Our approach improves visual memory metrics (SSIM) by 0.30 and PSNR by up to 10 dB over the baselines on this benchmark.
\end{itemize}

\section{Related Work}
\label{ch:rel}

\textbf{Video World Models.} Video world models simulate entire interactive 3D worlds via camera-/object-control \citep{nvidia2026cosmos3omnimodalworld,hyworld2026, robbyantteam2026advancingopensourceworldmodels, shen2026lyra, seo2026grounding, bruce_genie_2024, parkerholder2024genie2}, robotics simulators \citep{yang_learning_2024, zhu2025irasimfinegrainedworldmodel, quevedo2025worldgymworldmodelenvironment}, driving environments \citep{hu2023gaia1generativeworldmodel, russell2025gaia2controllablemultiviewgenerative, wang2023drivedreamerrealworlddrivenworldmodels, gao2024vistageneralizabledrivingworld, hassan2025gem}, games from user input \citep{valevski_diffusion_2025, che_gamegen-x_2024, zhang2025matrixgameinteractiveworldfoundation, he2026matrixgame20opensourcerealtime, qian2026matrixgame35enhancingrealtime, savov2025exploration}, while recent work extends this paradigm to desktop interfaces. ViMo \citep{luo_vimo_2025} addresses text corruption by separating textual and graphical generation through symbolic placeholders. NeuralOS \citep{rivard_neuralos_2025} uses a recurrent latent state to condition diffusion-based desktop generation. Neural Computers \citep{zhuge2026neuralcomputers} further explores learned software execution, identifying symbolic consistency and long-horizon memory as open challenges. These approaches improve interface simulation but lack explicit mechanisms for maintaining entity identity and tracking off-screen state changes. We maintain structured entity states and update alongside the generation.

\textbf{Memory-augmented world models.} FramePack \citep{zhang2025framecontextpackingdrift} compresses frame history, while Mixture of Contexts \citep{cai2025mixturecontextslongvideo} uses sparse attention routing for long-term retrieval. WorldMem \citep{xiao2025worldmemlongtermconsistentworld}, HyDRA \citep{chen2026out}, and VRAG \citep{chen2025learning} retrieve historical visual information to preserve consistency. Spatial memory further supports consistent revisits: VMem \citep{Li_2025_ICCV} indexes past views through surfels, SPMem \citep{wu2025videoworldmodelslongterm} maintains geometry-grounded memory, Captain Safari \citep{chou2026captainsafariworldengine} retrieves pose-aligned world tokens, and Lyra 2.0 \citep{shen2026lyra} uses geometry to retrieve and align historical observations. Beyond scene recall, WorldDirector \citep{wang2026worlddirector} preserves dynamic entities through LLM-planned trajectories specified before generation, while LiveWorld \citep{duan2026liveworld} simulates out-of-sight dynamics for selected entities, against a static 3D background. Our approach maintains an editable abstract state whose representation is not restricted to any specific environment.

\textbf{Language and Vision-Language Models.} Large language models support multi-step and compositional reasoning \citep{wei2023chainofthoughtpromptingelicitsreasoning, wang2023selfconsistencyimproveschainthought, zhou2023leasttomostpromptingenablescomplex, yao2023treethoughtsdeliberateproblem}, analogical inference \citep{webb2023emergentanalogicalreasoninglarge}, and representations of entity properties, relations, and spatiotemporal structure \citep{li-etal-2021-implicit, feng2024languagemodelsbindentities, gurnee2024languagemodelsrepresentspace}. Multimodal pretraining and instruction tuning extend these models to visual inputs \citep{alayrac2022flamingovisuallanguagemodel, li2023blip2bootstrappinglanguageimagepretraining, liu2023visualinstructiontuning, dai2023instructblipgeneralpurposevisionlanguagemodels, wang2024qwen2vlenhancingvisionlanguagemodels}. PaLM-E supports embodied reasoning \citep{pmlr-v202-driess23a}, ReAct interleaves reasoning and interaction \citep{yao2023reactsynergizingreasoningacting}, and RAP predicts action outcomes for planning \citep{hao2023reasoninglanguagemodelplanning}. Motivated by spatial reasoning in VLMs \citep{Chen_2024_CVPR}, we use one to predict updates to the abstract state.

\textbf{GUI data and benchmarks.} Agent benchmarks \citep{xie_osworld_2024, bonatti_windows_2024, zhou2024webarenarealisticwebenvironment, deng_mind2web_2023, rawles2024androidworld} and video corpora \citep{jian2026cuasuitemassivehumanannotatedvideo, chen_gui-world_2025} provide screenshots and action traces, but no persistent state identities linking actions to changes in previously visited views. GUI generation benchmarks evaluate image generators \citep{li2026gebenchbenchmarkingimagegeneration} or screenshot-level world models \citep{fu2026guiccbenchmarkingcontextualconsistency}, not frame-by-frame simulation with long-horizon revisits or state edits. We record aligned frames, actions, and state graphs, and test these capabilities.

\section{Method}
\label{ch:method}

In this section we first define the simulation problem and describe how \ours{} couples abstract-state prediction with video generation. We then present our data collection procedure and evaluation protocol for reconstruction, memory, and controllability.

\subsection{Problem setup}
\label{met:setting}
We model the environment as a partially observable controlled process with latent state $s_t = (G_t, v_t)$, comprising the full world state $G_t$ and a pointer $v_t$ to the focused node. The observation $o_t = \Omega(s_t) = G_t[v_t]$ contains only that node, while actions advance the state near-deterministically, $s_{t+1} = T(s_t, a_t)$. Transitions act on states rather than observations because actions can affect out-of-view elements. Frame generation follows $x_t \sim R(o_t, a_{t-1})$. Since the full state $s_t$ is not fully available in every environment, the simulator maintains an abstract state $M_t$ distilled from observations and updates it through sparse state updates, $M_{t+1} = M_t \oplus \Delta_t$. $M_0$ is the initial state representing the first observation alone. We keep the general structure but omit the rewards of the process, as our goal is simulation rather than planning.

\subsection{Model}
\label{met:model}

\begin{figure}[h]
\centering
\includegraphics[width=0.90\linewidth]{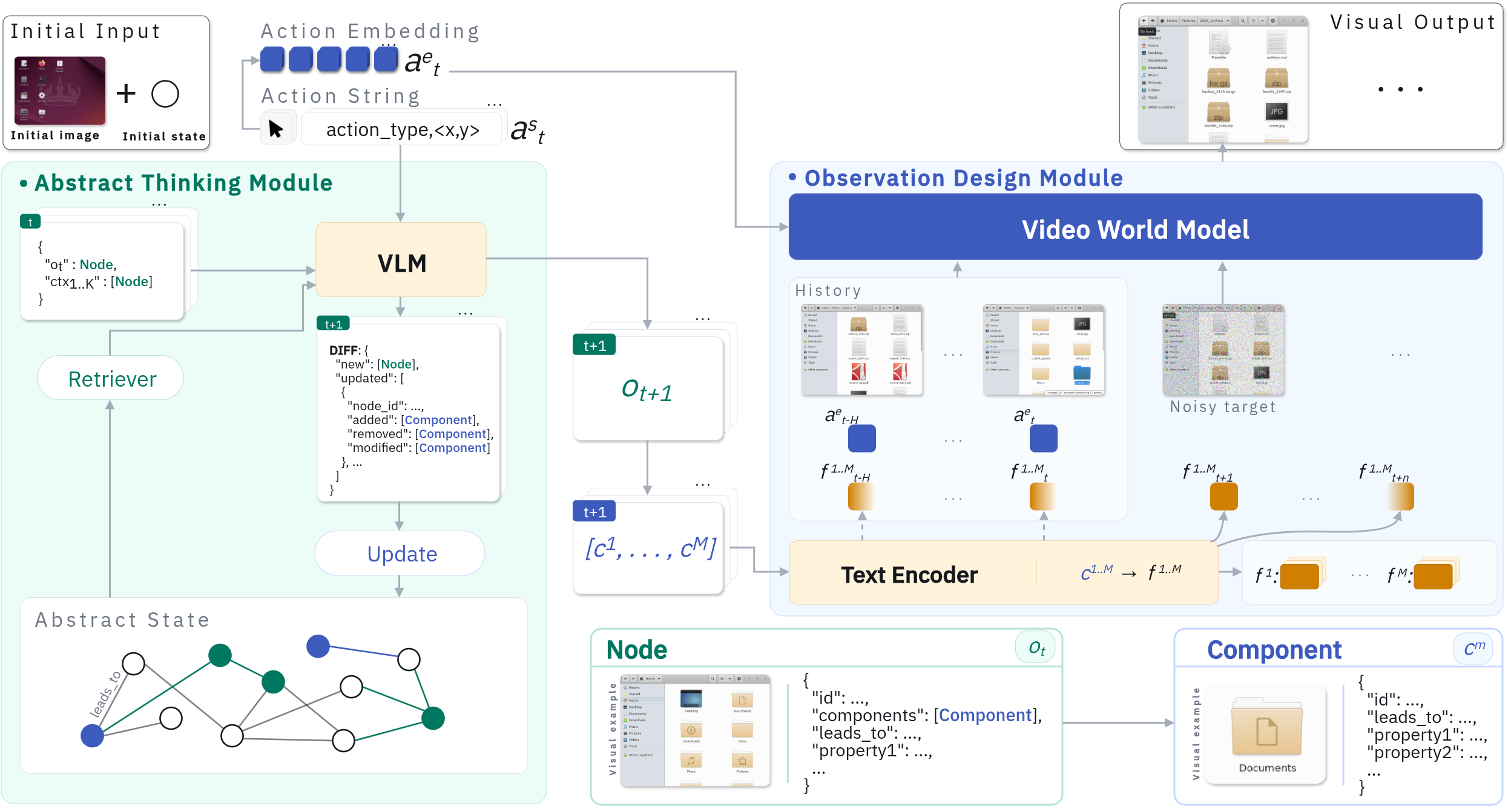}
\caption{\textbf{Overview of \ours{} for generative UI. 
} The abstract-thinking module (left) maintains the abstract state using a graph. Given action $a^s_t$, observed node $o_t$, and retrieved context nodes $\mathrm{ctx}_{1..K}$, a VLM predicts a state diff and the next node $o_{t+1}$. 
A frozen text encoder maps the predicted node's serialized components $c^{1..M}_{t+1}$ to embeddings $f^{1..M}_{t+1}$. The observation-design module (right) generates video frames conditioned on past frames, action embeddings $a^e_{t-H..t}$, and component embeddings $f^{1..M}_{t+1}$. Bottom-right panels detail the schema elements.}
\label{fig:architecture}
\end{figure}

\ours{} pairs two components operating in different spaces (Figure~\ref{fig:architecture}). Starting from an initial frame and its corresponding structured state, a VLM predicts action-induced updates to persistent abstract state, represented here as a graph, without receiving pixels. A diffusion model then generates the next frame conditioned on the predicted state, action, and visual history. The graph represents views and their constituent elements, preserving their content and spatial relationships across interactions.
 
One interaction step is factorized as
\begin{equation}
\small
p(x_{t+1}, o_{t+1}, \Delta_t \mid x_{\le t}, o_{\le t}, a_{\le t}) \approx
\underbrace{p_\phi(o_{t+1}, \Delta_t \mid o_t, a_t, M_t)}_{\text{abstract-space transition}}
\;\cdot\;
\underbrace{p_\theta(x_{t+1} \mid x_{\le t}, a_t, o_{t+1})}_{\text{video world model}},
\qquad M_{t+1}=M_t\oplus\Delta_t.
\label{eq:factorization}
\end{equation}

\subsubsection{Abstract state transition}
The transition model $p_\phi$ is a small pretrained VLM with low-rank finetuning \citep{hu2021loralowrankadaptationlarge}. Language models have been used to predict interface changes for selecting browser actions \citep{chae_web_2025}. Here, predicted changes update persistent abstract state and condition visual generation. The retriever resolves the destination from the current node, action, and targeted element, then retrieves the corresponding node if present in memory. For pointer actions, the targeted element is identified from the action coordinates and component geometry. Given the serialized current node, retrieved context, and action, the VLM predicts a transition decision (new, same, or revisited), the next node $o_{t+1}$, and a diff $\Delta_t$ that adds or modifies nodes and their components.

\subsubsection{Video world model}
The video world model $p_\theta$ generates each frame conditioned on previous frames, the current action, and the predicted next abstract state. The action is supplied as text to the state predictor and as an embedding to the video model. The approach supports diffusion backbones with cross-attention conditioning. Because text and changing elements occupy small image regions, the training loss assigns them higher weight (Equation~\ref{eq:loss}). Implementation and training details are in Appendix~\ref{app:arch}.

\subsubsection{Abstract state conditioning}
The predicted node $o_{t+1}$ is serialized into component strings, each independently embedded by a frozen text encoder. The resulting features condition the corresponding frame through cross-attention after each transformer block's self-attention \citep{zhuge2026neuralcomputers}:
\begin{equation}
H^k \leftarrow H^k + W_o\,
\mathrm{softmax}\!\left(\tfrac{(H^kW_q)(F^kW_k)^\top}{\sqrt{d}}\right)
F^kW_v,
\label{eq:crossattn}
\end{equation}
where $H^k$ contains the visual tokens of frame $k$, $F^k$ its state-component embeddings, $W_q,W_k,W_v,W_o$ are learned projections, and $d$ is the query/key dimension. Each frame attends only to its corresponding state features, maintaining alignment between predicted state and visual generation. Implementation details are in Appendix~\ref{app:stateCond}.

\begin{figure}[h]
\centering
\includegraphics[width=0.85\linewidth]{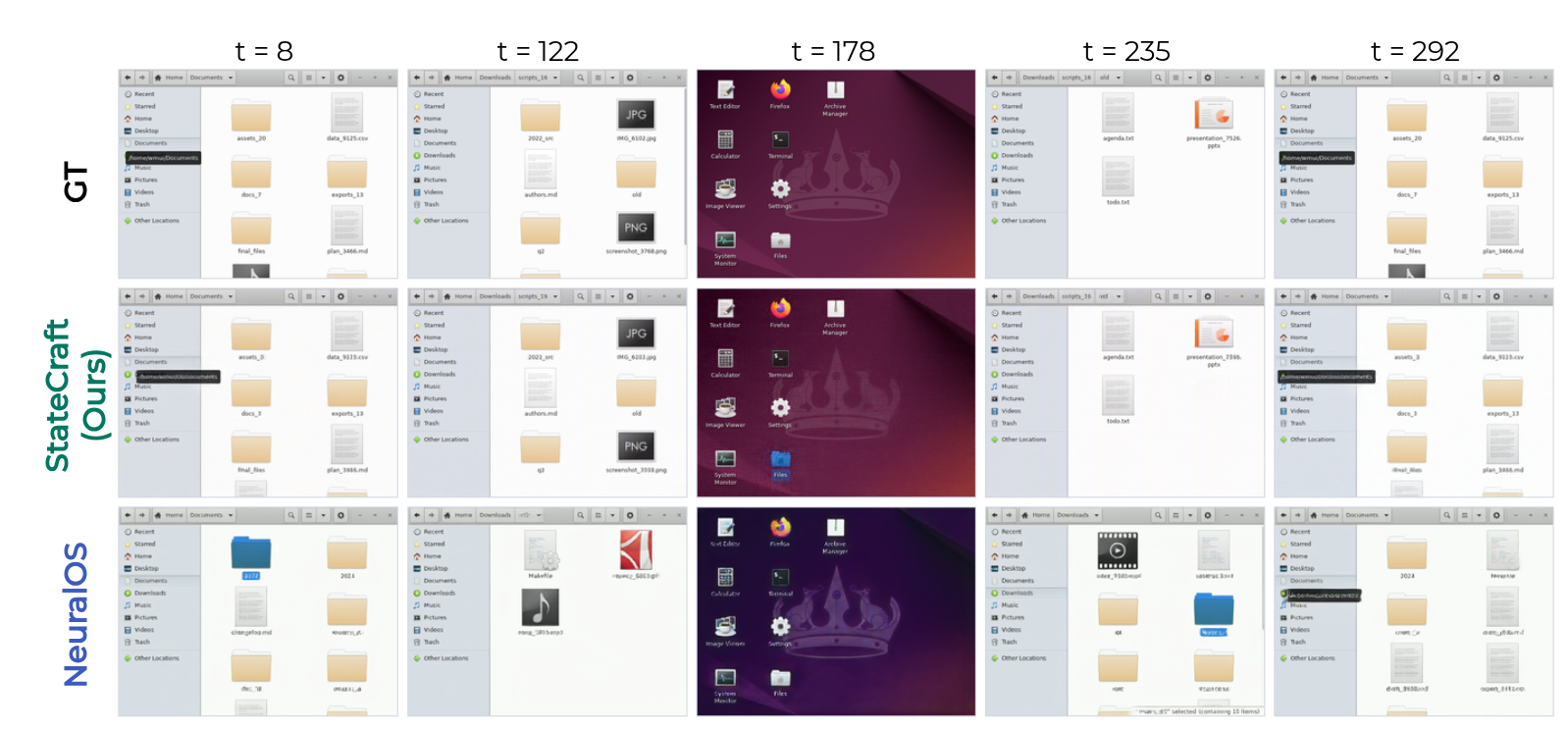}
\caption{\textbf{Autoregressive rollout on a held-out trajectory.} The model conditioned on the state signal tracks folders, filenames, and icons across the whole rollout, while NeuralOS keeps a convincing shell but drifts in content and loses high-frequency details.}
\label{fig:longRollout}
\vspace{-5mm}
\end{figure}

\subsection{Data collection}
\label{met:data}

We collect trajectories in GNOME Files on containerized Ubuntu with randomly generated filesystems. Instead of sampling inputs blindly \citep{zhuge2026neuralcomputers,rivard_neuralos_2025}, our policy selects actions on interactable elements using the accessibility tree, which describes interface elements and their properties. The training set contains 1,956 trajectories and 586,210 actions, of which 47.7\% lead to a visible change in the state. For evaluation, we collect 240 held-out trajectories of 300 actions. For memory, we construct 184 navigation sequences that enter a target folder, visit other folders, and return. Each is repeated at five revisit horizons, keeping target contents and revisit frames identical (920 trajectories). For controllability, we construct 293 clips from held-out trajectories by replacing element names or types, or inserting elements into the conditioning state and inpainting the result in the corresponding frame, while keeping the preceding visual context and action sequence unchanged. Using AT-SPI \citep{atspi2}, we record aligned images, actions, and observations as $\tau=(x_0,o_0,(a_t,x_t,o_t)_{t=1}^{L})$. Graph nodes represent views identified by folder path or dialog type, allowing repeated visits to reference the same node. We store the initial graph and subsequent changes, reset the environment between trajectories, and discard discontinuous recordings. Appendix~\ref{app:data} provides terminology, collection details, and dataset splits.

\subsection{Benchmark}
\label{met:metrics}

We introduce a protocol evaluating the state consistency required of UI world models. Full-frame PSNR and SSIM \citep{wang2004image} can mask content errors because most pixels belong to static interface elements. Aggregate scores also overrepresent frequently visited views, such as home screens. We therefore compute crop-level metrics over state-bearing regions (path, main pane, and side list), using recorded element bounds and OCR-localized text boxes \citep{du2020ppocrpracticalultralightweight}, and weight frames by inverse frequency of the state and element appearance.

\subsubsection{Memory, navigation, and controllability}
Memory is measured as self-consistency: at each ground-truth revisit, we compare the model's generated frame with its own frame from the first visit, using ROI-level SSIM with inverse-frequency pooling over disjoint lag buckets. Scores lie in $[0,1]$, plausible but unrelated folders score approximately 0.5.

For commanded edits, the contrastive follow score indicates whether the prediction is closer to the post-edit than the pre-edit appearance over changed pixels. Let NEW and OLD denote these ground-truth appearances, and $d_{\mathrm{new}}$ and $d_{\mathrm{old}}$ their respective mean absolute distances from the prediction within the change mask $\mathcal{M}$. Then
$\text{follow}=\mathbf{1}\!\left[\, d_{\mathrm{new}} < d_{\mathrm{old}} \,\right].$
Reproducing the pre-edit state scores zero. We additionally report PSNR within the edited box and at an unedited control element.

\section{Experiments}
\label{exp:general}

Unless stated otherwise, models are trained on the same corpus of $\sim$2000 trajectories of 300 actions each. To assess whether abstract-state conditioning benefits different generation targets, we implement \ours{} with pixel-space DFoT \citep{song_history-guided_2025} and latent-space CF++ \citep{zhu2026causal,zhao2026causal}, based on pretrained Wan2.1 \citep{wan2025}. Our external baseline is NeuralOS \citep{rivard_neuralos_2025}, a latent diffusion desktop simulator with recurrent memory, trained on the same corpus following its public schedule and iteration count. All models are evaluated at $256\times192$, downsampling higher-resolution outputs. Backbone configurations and training details are in Appendix~\ref{app:arch}.

\subsection{Data collection evaluation}
\label{exp:data}
We compare grounded collection with blind input sampling \citep{rivard_neuralos_2025,zhuge2026neuralcomputers}, using regenerated NeuralOS-style data and a blind collector in our environment. Grounded collection produces three times as many transitions with substantial visual changes as blind collection in the same environment (63.9\% versus 21.3\%) and reaches deeper folders more frequently, while visiting approximately double the amount of unique state over 300 actions. These results support targeting interactable elements to collect informative training trajectories. Corpus descriptions, statistical analyses, and coverage estimates are provided in Appendix~\ref{app:data-richness}.

\subsection{Generation under long rollouts}
\label{exp:consist}

We evaluate on 240 held-out file-manager trajectories of 300 actions each. To assess state evolution and retention without requiring models to guess arbitrary unseen filenames, we supply first-visit filenames to \ours{} as element hints and to CF++-based baselines through text prompts. NeuralOS receives the corresponding ground-truth frame to initialize its recurrent memory. No additional target information is supplied on revisits. Generation proceeds autoregressively between initialization events. All methods are scored at identical frame indices, excluding the indices corresponding to first visits of directories.
Abstract-state conditioning improves both backbones across all metrics (Table~\ref{tab:diffGeneralMetrics}). For CF++, it increases PSNR by 10.29 dB over the unconditioned model and comes within 0.45 dB of ground-truth conditioning. Gains are largest on the canvas: text-region SSIM reaches 0.85, compared with 0.16 for CF++ and 0.15 for NeuralOS (Table~\ref{tab:elemMetrics}), indicating improved reconstruction of variable interface content. Figure~\ref{fig:longRollout} illustrates this over a rollout. Subsequent experiments use CF++, our best-performing backbone. Appendix~\ref{app:dfot} reports DFoT reconstruction scores by screen region and within text boxes.

\begin{table}[t]
\centering
\small
\caption{\textbf{Generation quality with first-visit initialization.} Our method improves every metric over the unconditioned backbones and NeuralOS, and stays close to the reconstruction quality obtained with ground-truth state updates.}
\label{tab:diffGeneralMetrics}
\begin{tabular}{lcccc}
\toprule
\textbf{Model}
& \textbf{SSIM} $\uparrow$
& \textbf{PSNR} $\uparrow$
& \textbf{ifreq-SSIM} $\uparrow$
& \textbf{ifreq-PSNR} $\uparrow$ \\
\midrule
NeuralOS & 0.85 & 20.92 & 0.82 & 19.04  \\
DFoT & 0.86 & 23.27 & 0.81 & 19.45 \\
CF++ & 0.78 & 18.49 & 0.79 & 18.99 \\
\ours{} (Ours; DFoT) & \underline{0.92} & \underline{27.63} & \underline{0.92} & \underline{28.06} \\
\ours{} (Ours; CF++) & \textbf{0.95} & \textbf{28.78} & \textbf{0.95} & \textbf{28.71} \\
\multicolumn{5}{l}{\footnotesize\itshape Upper bound: ground-truth conditioning} \\
\midrule
\textit{\ours{} (Ours; DFoT)} & 0.93 & 29.55 & 0.95 & 31.85 \\
\textit{\ours{} (Ours; CF++)} & 0.95 & 29.23 & 0.96 & 30.36 \\
\bottomrule
\end{tabular}
\vspace{-2mm}
\end{table}

\begin{table}[t]
\centering
\small
\caption{\textbf{Render quality by screen region, over the full region (top) and inside OCR-detected text boxes only (bottom).} Gains over the baselines are largest on the canvas, the region with the highest variability. Side panel is mostly constant and the models do not need external conditioning to perform well. Last row of each block: upper bound with ground-truth diffs.}
\label{tab:elemMetrics}
\begin{tabular}{lcccccc}
\toprule
& \multicolumn{2}{c}{\textbf{Breadcrumb}} & \multicolumn{2}{c}{\textbf{Canvas}} & \multicolumn{2}{c}{\textbf{Side panel}} \\
\cmidrule(lr){2-3} \cmidrule(lr){4-5} \cmidrule(lr){6-7}
\textbf{Model} & SSIM$\uparrow$ & PSNR$\uparrow$ & SSIM$\uparrow$ & PSNR$\uparrow$ & SSIM$\uparrow$ & PSNR$\uparrow$ \\
\midrule
\multicolumn{7}{l}{\footnotesize\itshape Full region} \\
NeuralOS & \underline{0.85} & \underline{30.12} & \underline{0.61} & \underline{19.74} & \textbf{0.96} & \textbf{34.28} \\
CF++ & 0.77 & 21.86 & 0.59 & 17.51 & 0.88 & 23.99 \\
\ours{} (Ours; CF++) & \textbf{0.91} & \textbf{32.01} & \textbf{0.89} & \textbf{30.80} & \underline{0.95} & \underline{33.06} \\
\textit{\ours{} (Ours; CF++), GT diffs} & 0.93 & 32.20 & 0.94 & 32.35 & 0.95 & 33.53 \\
\midrule
\multicolumn{7}{l}{\footnotesize\itshape OCR-detected text boxes} \\
NeuralOS & \underline{0.52} & \underline{27.16} & 0.15 & \underline{16.71} & \underline{0.80} & \underline{27.90} \\
CF++ & 0.38 & 20.46 & \underline{0.16} & 15.88 & 0.42 & 20.01 \\
\ours{} (Ours; CF++) & \textbf{0.76} & \textbf{30.46} & \textbf{0.85} & \textbf{25.86} & \textbf{0.88} & \textbf{28.71} \\
\textit{\ours{} (Ours; CF++), GT diffs} & 0.80 & 30.48 & 0.90 & 27.08 & 0.89 & 28.73 \\
\bottomrule
\end{tabular}
\vspace{-5mm}
\end{table}

\subsection{Memory evaluation}
\label{exp:mem}

\begin{figure}[t]
\centering
\begin{subfigure}[t]{0.35\linewidth}
\centering
\includegraphics[width=\linewidth]{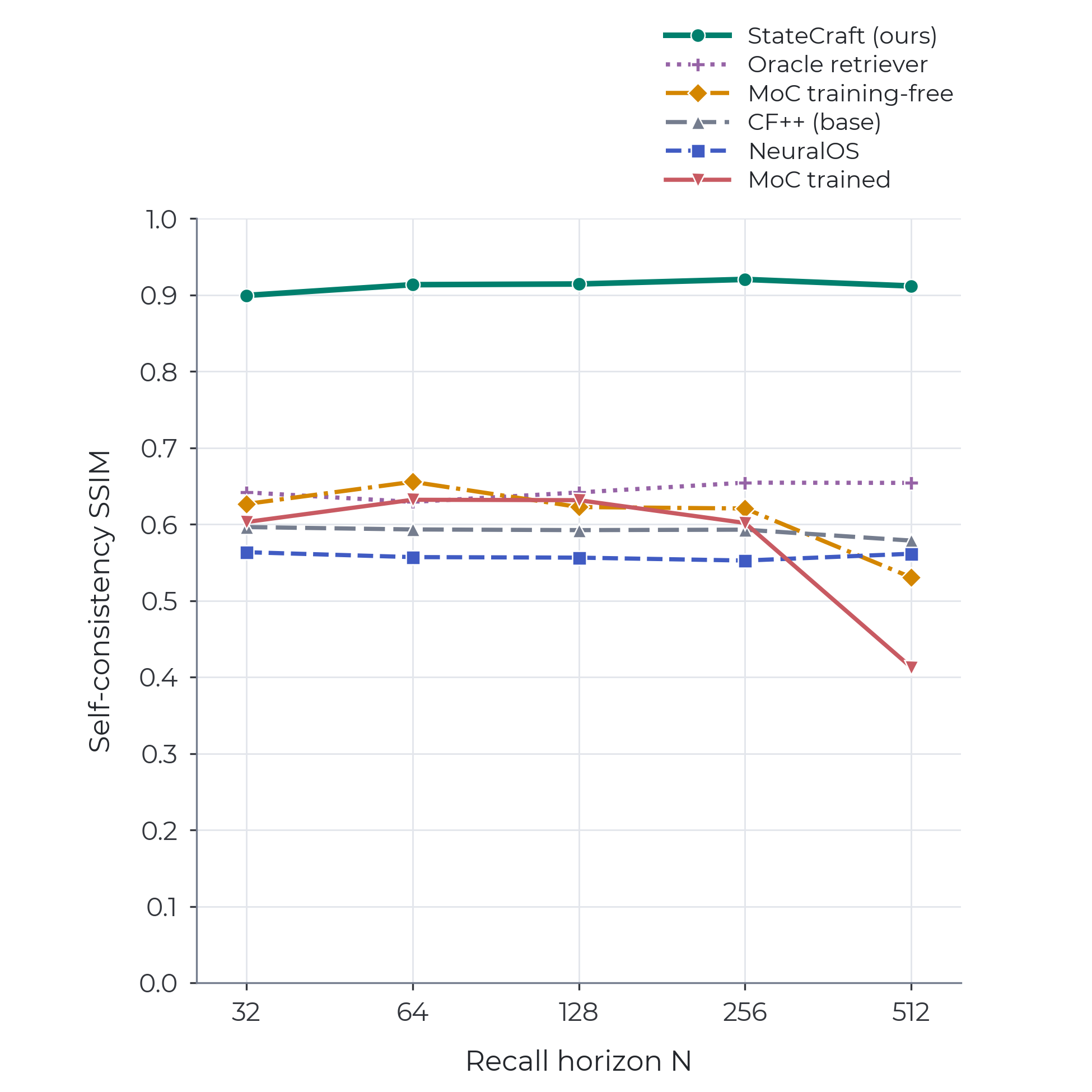}
\caption{Revisit self-consistency SSIM versus horizon $N$.}
\label{fig:memRecall}
\end{subfigure}
\hfill
\begin{subfigure}[t]{0.6\linewidth}
\centering
\includegraphics[width=\linewidth]{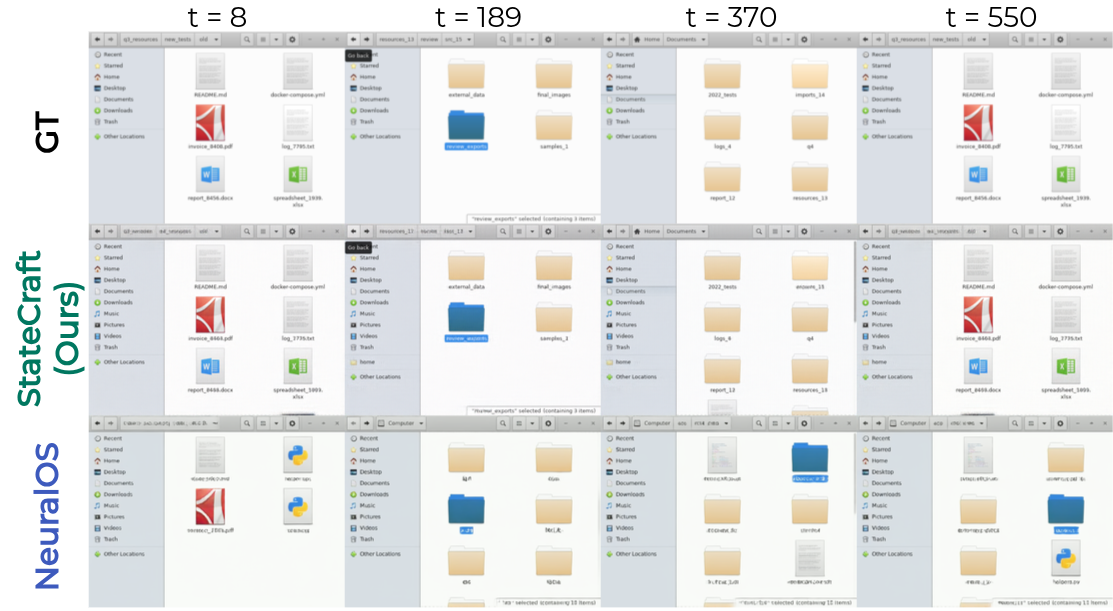}
\caption{Target reconstruction at $N{=}512$.}
\label{fig:memRevisit}
\end{subfigure}
\caption{\textbf{Memory across revisit horizons.}
\textbf{(a)} \ours{} maintains high self-consistency across horizons. Most baselines score around 0.5--0.6, trained MoC declines at the longest horizons.
\textbf{(b)} \ours{} preserves target content, whereas NeuralOS generates a plausible interface with hallucinated content.}
\label{fig:memory}
\end{figure}

We evaluate 184 matched trajectory families, each entering a deep target folder, traversing distractors, and returning after $N \in \{32,64,128,256,512\}$ actions, yielding 920 trajectories. Target contents remain unchanged, and target-visit subsequences are pixel-identical across horizons (zero MSE), controlling appearance while varying rollout length. All models receive the same eight ground-truth context frames, including the first target observation, and generate two further target frames before distraction. Corresponding filenames are supplied to \ours{} as element hints consistent with the visual context and to CF++-based baselines through text prompts. No target content is supplied again on revisits. We score only the designated target using ROI-level SSIM between generated first-visit and revisited frames (Section~\ref{met:metrics}). 
This revisit target SSIM measures self-consistency rather than correctness and complements the reconstruction metrics in Section~\ref{exp:consist}. 
Figure~\ref{fig:memory} shows that \ours{} maintains high revisit similarity across horizons and substantially outperforms the baselines.
Appendix~\ref{app:mem} details initialization and pairing and reports target-revisit similarity to ground truth across the same horizons, measuring reconstruction correctness. We also evaluate training-free and finetuned adaptations of Mixture of Contexts \citep{cai2025mixturecontextslongvideo}, and an oracle-retrieval of past latents based on GT state identities unavailable at deployment. Both are based on CF++. Implementation and training details are in Appendices~\ref{app:moc} and~\ref{app:oracle}.

\subsection{Controllable generation}
\label{exp:ctrl}

Our architecture allows direct edits to persistent abstract state. We test whether generation follows these edits when they conflict with visual context, using 293 held-out clips (Section~\ref{met:data}). We modify one or more canvas elements in the next frame's conditioning, keeping visual context, actions, and other tokens unchanged. Edits include label and category substitutions, multi-element substitutions, and insertions, under ground-truth or self-generated context and when entering or revisiting a directory. The baseline uses frames generated by the same model without the intervention. Table~\ref{tab:ctrlBuckets} shows an overall edit-following rate of 81\%, compared with 0\% without intervention, reaching 98\% for category changes upon re-entry but 48\% for insertion. Edited-region PSNR improves from 17.6 to 27.62 dB, although control-region PSNR decreases from 37.7 to 23.88 dB. Thus, state edits guide generation beyond visual history, with limitations in inserting new elements and preserving unedited regions. Figure~\ref{fig:ctrl} in Appendix~\ref{app:vis} provides visual examples.

\begin{table}[t]
\centering\small
\caption{\textbf{Controllability by edit type and context.} Edit-following rate and PSNR within edited (E) and unedited control (C) crops. The baseline uses the same model without the state intervention. All predictions are evaluated against the same edited targets using the metrics of Section~\ref{met:metrics}.}
\label{tab:ctrlBuckets}
\begin{tabular}{llccc}
\toprule
\textbf{Edit} & \textbf{Context at the edited frame}
& \textbf{Follow} & \textbf{E-PSNR} & \textbf{C-PSNR} \\
\midrule
\multicolumn{5}{l}{\textit{Label change} (same file type: new name with icon fixed)} \\
\quad Rename & ground truth                     & 0.64 & 25.1 & 21.3 \\
\quad Rename & model-generated context           & 0.92 & 30.9 & 22.1 \\
\addlinespace
\multicolumn{5}{l}{\textit{Type change} (new file type and name)} \\
\quad Retype & ground truth                    & 0.84 & 28.4 & 22.6 \\
\quad Retype & model-generated context           & 0.86 & 26.9 & 24.6 \\
\quad Retype & directory entered one frame ago     & 0.83 & 27.5 & 24.9 \\
\quad Retype & directory re-entered at same step (hidden change) & 0.98 & 32.2 & 25.7 \\
\quad Multi-retype & ground truth       & 0.83 & 26.6 & 25.4 \\
\addlinespace
\multicolumn{5}{l}{\textit{Insertion} (element added on empty space)} \\
\quad Insert & settled view                     & 0.48 & 20.8 & 27.5 \\
\midrule
\multicolumn{2}{l}{\textit{Baseline} (unedited generated frame)} & 0.00 & 17.6 & 37.7 \\
\midrule
\textbf{All} & & \textbf{0.81} & \textbf{27.62} & \textbf{23.88} \\
\bottomrule
\end{tabular}
\vspace{-5mm}
\end{table}

\subsection{Ablation Study}
\label{exp:ablation}

Evaluated independently, the finetuned state predictor achieves 0.86 node and 0.96 transition accuracy on held-out navigation, versus 0.48 and 0.52 for the few-shot base model, showing that lightweight finetuning learns state dynamics. For our model of choice, this comes at the cost of out-of-domain generality (Table~\ref{tab:llmEval}). Ablations on DFoT and CF++ show complementary conditioning roles: removing graph tokens reduces canvas SSIM by 0.31 on average and revisit SSIM by over 0.20, while removing actions causes smaller reconstruction losses. For DFoT, path-only conditioning barely improves canvas or memory scores. Encoding the full node helps, while separate element tokens further raise canvas SSIM from 0.76 to 0.92 (Appendices~\ref{app:llm} and \ref{app:cond-ablation}).

\subsection{Event-driven generation}
\label{exp:probe}

Following work on probing internal world representations \citep{li2023emergent, esmati2026invisiblehandphysicsvideo}, we train small classifiers on frozen video-model activations, supervised by recorded ground-truth states, to detect changes and distinguish five transition types, including no change, backward navigation, forward navigation, window closure, and navigation to a nonadjacent folder. Binary and five-class probes achieve AUROC above 0.97 (Appendix~\ref{app:probe-setup}). After generating frame $x_t$, the gate combines cached activations with incoming action $a_t$ to decide whether to invoke the language model. The updated or retained state conditions $x_{t+1}$, requiring no additional backbone pass. In closed-loop evaluation, the gate skips 40.8\% of language-model calls with 91.2\% skip purity, reducing PSNR by 0.57 dB and SSIM by 0.008. This enables reciprocal interaction where abstract state conditions visual generation, while visual features govern subsequent state updates.

\begin{figure}[t]
\centering
\includegraphics[width=0.9\linewidth]{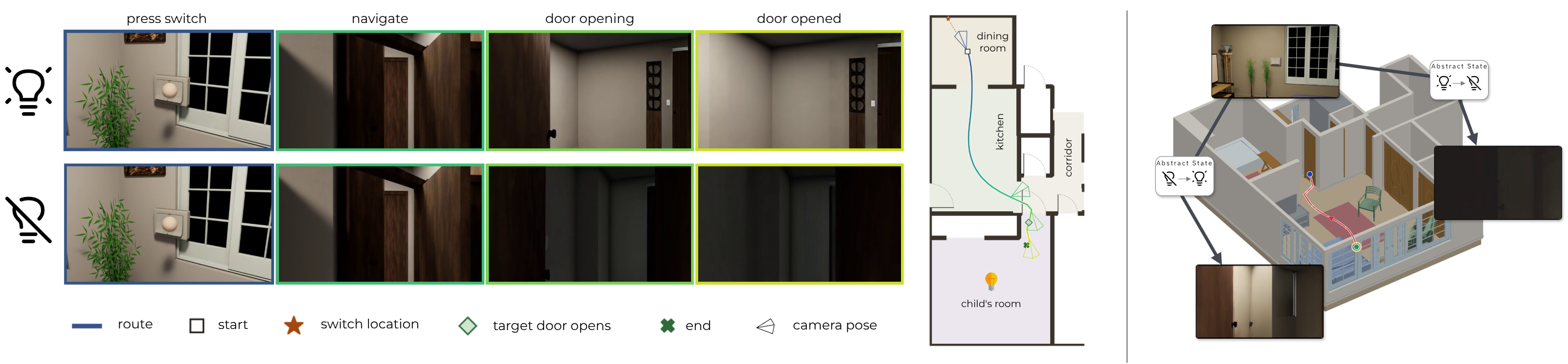}
\caption{\textbf{Hidden-state conditioned generation.} Left: two rollouts along the same path with different state signals. Our model generates the corresponding target-room illumination (last column) while preserving the distractor-room state (second column). The floor plan marks the camera path and four rendered poses. Right: another example from the second held-out apartment.}
\label{fig:house}
\vspace{-5mm}
\end{figure}

\subsection{Hidden-state conditioning beyond software simulation}
\label{exp:hidden_3d}

As a proof of concept beyond interfaces, we test abstract-state conditioning in simulated 3D interiors \citep{li2024behavior1khumancenteredembodiedai}, where a switch controls an unobserved room's illumination. Room category and illumination state condition the video model, isolating its ability to render off-screen action effects. Trained on 8,644 trajectories across 14 houses, our model achieves 0.864 illumination-classification AUC on 120 balanced episodes from two unseen houses, versus 0.542 for separately trained CF++ without state conditioning. With visual context and noise fixed, flipping the illumination token shifts mean brightness by 63.7\% of the ground-truth difference ($p=0.0006$ across seven scene groups; Figure~\ref{fig:house}). These results support state-conditioned generation beyond interfaces, including unobserved action effects. Appendix~\ref{app:3dexp} provides experimental details.

\section{Conclusion}
\label{ch:conclusion}
We introduce a paradigm that decouples abstract-state reasoning from visual generation. \ours{} pairs a language model that updates persistent state with a video diffusion model that renders it. Our benchmark demonstrates how abstract-state reasoning supports visual quality, long-term consistency, and controllability in generative interfaces. A 3D proof of concept extends state-conditioned generation to spatial environments. These results support explicit, evolving memory as a foundation for consistent and controllable world models.

\textbf{Limitations.} Our approach relies on a predefined state schema for each environment. Learning the schema and acquiring new content autonomously remain future work. Broader evaluation is needed across applications, changing environments, and 3D settings with learned state prediction. Improving robustness to accumulated state errors and establishing end-to-end latency gains are further directions.

\bibliographystyle{unsrtnat}
\bibliography{references}

\newpage
\appendix
\section{Appendix}
\label{app}

\etocsettocstyle{\subsection*{Table of Contents}}{}
\etocsetnexttocdepth{subsubsection}
\localtableofcontents

\subsection{Model}
\label{app:model}

\subsubsection{State conditioning implementation}
\label{app:stateCond}
We serialize each node into at most $M=32$ component strings and embed them independently using a frozen Qwen3 text encoder \citep{zhang2025qwen3embeddingadvancingtext}. Embeddings are truncated to 1024 dimensions using its Matryoshka representation \citep{kusupati2022matryoshka}. Cross-attention output projections are zero-initialized. Training begins with the video backbone frozen, followed by joint finetuning of the backbone and conditioning layers; the text encoder remains frozen throughout.

\subsubsection{Diffusion architecture and training details}
\label{app:arch}

\paragraph{Backbones.}
The pixel-space denoising network is a residual U-ViT after Simple Diffusion
\citep{hoogeboom2023simplediffusionendtoenddiffusion}: a $2\times2$
convolution patchifies each frame, which passes through a four-level
hierarchy whose channel width climbs from 128 to 1024. The two finest
resolutions use convolutional residual blocks; the two coarsest use
transformer blocks, three per path at the intermediate level and sixteen at
the bottleneck, with a single residual skip at each resolution. At a given resolution the tokens of all 32 frames of a training clip form one joint sequence, on the order
of twenty-five thousand tokens at the intermediate level and six thousand at
the bottleneck. Each block runs attention and feed-forward in parallel with
query-key normalisation, and initialises every output projection to zero.
Per-frame conditioning (noise-level embedding plus action embedding) reaches
every block through feature-wise scale-and-shift modulation
\citep{perez2017filmvisualreasoninggeneral}. The network totals roughly 364M
parameters: 288M in the action-conditioned base model, 76M in the
cross-attention modules added for graph conditioning.

The causal-forcing backbone (CF++) starts from the Wan2.1-1.3B diffusion
transformer \citep{wan2025} and its VAE, finetuned under the causal forcing
recipe \citep{zhu2026causal, zhao2026causal} so that generation is frame-autoregressive:
each frame is denoised in 2 steps conditioned only on past frames, at
$640\times480$. Action and graph conditioning enter both through
cross-attention, following the ideas presented in \cite{zhuge2026neuralcomputers}, with output projections zero-initialized so the pretrained model is
undisturbed at insertion. This version has a total of 1.89B parameters: 1.42B in the Wan2.1 backbone and 470M in the added conditioning cross-attention layers. Actions use the encoding described below, projected to cross-attention tokens. Because the model denoises in latent
space, the per-pixel weight map of Equation~\ref{eq:loss} is projected onto the
latent grid at the VAE's downsampling factor.

\paragraph{DFoT objective and sampling.}
Training of the pixel-space backbone is formulated in continuous time under
diffusion forcing: each frame is assigned an independent diffusion time
$t_k \sim \mathcal{U}[0,1]$ and corrupted as
$x^k_{t_k} = \alpha_{t_k} x^k + \sigma_{t_k} \varepsilon^k$ with
$\varepsilon^k \sim \mathcal{N}(0, I)$, the scales derived from the log
signal-to-noise ratio under a resolution-shifted cosine schedule
\citep{hoogeboom2023simplediffusionendtoenddiffusion}. The network predicts
the velocity target \citep{salimans2022progressivedistillationfastsampling}
and minimizes the equivalent noise error
\begin{equation}
\mathcal{L}(\theta) =
\mathbb{E}\left[ \sum_{k} w(\lambda_{t_k})
\left\| \hat{\varepsilon}^{\,k} - \varepsilon^k \right\|^2_{m^k} \right],
\label{eq:loss}
\end{equation}
with the sigmoid weighting of
\citep{kingma2023understandingdiffusionobjectiveselbo} and the per-pixel
weight map $m^k$ built from the recorded element bounds and an offline
text-detection pass, assigning weight $\times20$ to dynamic elements, $\times50$ to text, and 1 elsewhere. Attention in the transformer stages is joint over space
and time with axial 3D rotary position
\citep{su2023roformerenhancedtransformerrotary}. The action embedding, a
verb embedding concatenated with a smooth sinusoidal encoding of the click
coordinates, is dropped with probability 0.1 after the manner of
classifier-free guidance \citep{ho2022classifierfreediffusionguidance},
independently of the graph channel, which is dropped with probability 10\%. Sampling for the pixel-space
backbone is deterministic DDIM
\citep{song2022denoisingdiffusionimplicitmodels} over 50 steps; rollout
slides a window keeping the eight most recent frames as context, so after the first eight ground-truth frames the model conditions only on its own outputs.

\paragraph{CF++ objective and sampling.}
CF++ adapts Wan2.1-T2V-1.3B to frame-by-frame generation with the causal forcing recipe \citep{zhu2026causal, zhao2026causal}. Starting from the public weights, we warm up the graph cross-attention for 30k steps, train the causal diffusion stage for 39k steps, and evaluate the final consistency-distillation stage, where we apply the latent-projected weight map. The VAE encodes each frame separately, so actions and graph tokens align one-to-one with latent frames, with frame $t{+}1$ receiving the transition of frame $t$. We train on 21-frame clips; at inference, each frame is generated in 2 denoising steps with a key-value cache of up to 21 previous frames, after 8 ground-truth frames. Wan's text input receives the ground-truth path and first-visit filenames of each folder for the CF++ baselines (Section~\ref{exp:consist}), and the fixed prompt \texttt{"desktop file manager UI transition"} for \ours{}.

\paragraph{Action encoding.}
The interaction verb, one of six classes spanning single, double, and right
clicks, menu activations, window-management operations, and the null action,
is a learnable embedding. Click coordinates, normalized to the unit square,
are polar-decomposed, lifted onto the unit three-sphere, and unrolled into
sinusoidal features at logarithmically spaced frequencies, giving a
positional encoding that varies smoothly across the screen; verb and position
encodings are concatenated and projected to the width of the modulation
pathway.

\paragraph{Training schedule.}
For DFoT, the action-conditioned base model trains first, alone. The cross-attention
modules are then inserted and trained with the rest of the network frozen,
their zero initialization leaving the base behavior intact; this warm-up
shields the pretrained dynamics from the gradient noise of fresh modules and
lets the attention patterns onto the graph tokens take shape before the
backbone co-adapts. The whole network is then finetuned jointly. An
exponential moving average of the weights is kept throughout and used for
every reported result. Training runs to 110k iterations (roughly 400 H200 GPU hours). 
CF++ follows the same staged schedule, starting from pretrained Wan2.1: 40k iterations of initial finetuning, 30k iterations of warm-up and 39k of joint finetuning plus a final 1k of consistency distillation(roughly 400 H200 GPU hours).
Conditioning variants (described in Appendix~\ref{app:cond-ablation}) follow a budget-matched protocol: all start from the same
action-conditioned checkpoint at 40k iterations, graph-conditioned variants are
finetuned a further 70k iterations (30k conditioning warm-up, 40k joint finetuning), and the
unconditioned baseline trains for the same added budget, so no variant sees
more updates than another. The second stage enables the region-of-interest
reweighting of Equation~\ref{eq:loss} for all variants.

\begin{table}[h]
\centering\small
\caption{Training hyperparameters.}
\label{tab:hparams}
\begin{tabular}{lcc}
\toprule
& DFoT & CF++ \\
\midrule
Parameters (base / added)     & 288M / 76M          & 1.42B / 470M \\
Training clip length (frames) & 32                  & 21 \\
Context at inference (frames) & 8                   & 21 \\
Denoising steps               & 50 (DDIM)           & 2 \\
Optimizer / learning rate     & AdamW / $1\times10^{-4}$ & AdamW / $2\times10^{-6}$ \\
Batch size                    & 4 $\times$ 2 accum $\times$ 4 GPU = 32 & 8 (1 per GPU $\times$ 8) \\
Iterations (initial / warm-up / joint)  & 40k / 30k / 40k           & 40k / 30k / 39k + 1k CD \\
EMA decay                     & 0.9999              & 0.99 \\
Compute                       & $\sim$400 H200 h    & $\sim$400 H200 h \\
\bottomrule
\end{tabular}
\end{table}

\paragraph{NeuralOS.}
We train NeuralOS \citep{rivard_neuralos_2025} with its public code and schedule at its native $512\times384$ resolution, for 110k iterations to match our budget, batch size 32. NeuralOS training is also split in 3 stages (RNN training, diffusion training, joint finetuning), so we map our 40k/30k/40k splits exactly to their architecture. NeuralOS training takes approximately 250 H200 GPU hours.

\subsubsection{State-predictor training}
\label{app:llm-training}

We use Qwen3.5-0.8B \citep{qwen35blog}, adapted through low-rank finetuning \citep{hu2021loralowrankadaptationlarge} on recorded state transitions. We use LoRA with rank 16, applied to all attention and MLP projections (q, k, v, o, gate, up, down). We train with an effective batch of 32 (8 on a single H200 $\times$ 4 accumulation steps) at 16384 max sequence length, learning rate $2\times10^{-4}$, decay at 0.01, for 12k steps of which 100 are warmup. We select the checkpoint with the best structural validation composite, evaluated every 200 steps on a fixed 1,024-example held-out slice, and stop when neither that composite nor validation loss improves for 10 consecutive evaluations.

\paragraph{Inputs and leakage prevention.}
Inputs are kept from leaking answers: internal node identifiers and oracle
transition labels are removed, and the identity of the clicked element is
not handed over but re-derived from the click coordinates against the
current node's component geometry, so that the model must ground the action
spatially just as its downstream consumer does.

\paragraph{Objective and model selection.}
Finetuning minimizes the next-token cross-entropy on the completion alone,
with two adjustments. First, the loss is reweighted across output sections,
leaning on the transition decision and the predicted paths over the long
tail of component text, and supervision on rarely seen filename tokens is
stochastically down-weighted; without this the model memorizes the training
name distribution instead of learning the shape of the transition. Second,
model selection rests not on the scalar validation loss, which the
memorization failure mode can drive down even as the useful capability
degrades, but on a structural score that multiplies the rate of well-formed
outputs by per-field quality measures, path correctness, component-kind
agreement, and bounding-box placement, kept deliberately blind to the exact
filename. Training examples are rebalanced across transition types, holding
on to every state-changing and revisit step while thinning the uneventful
ones. 

\paragraph{Retrieval and decoding.}
During training, the memory retrieval that surfaces revisited nodes comes
from an oracle on half the trajectories and from the same rule-based address
resolution used at inference on the other half, so the model never trains on
a retrieval distribution it will not meet once deployed. At inference it
decodes greedily. We explored the option of having the model operate under a grammar constraint that limits
generation to the legal completion shapes and to reference paths actually present in the prompt, but we found it overall detrimental to performance compared to letting the model generate freely.

\subsection{Data Collection Framework}
\label{app:data}

\paragraph{Interface terminology.}
An accessibility tree describes interface elements through their roles, labels, and screen bounds; AT-SPI provides access to this information. In the file manager, the breadcrumb displays the current path, the main pane displays folder contents, and the sidebar provides navigation shortcuts. Our state graph records views as nodes and individual elements as their components.

\subsubsection{Collection and state recording}
\label{app:collection-implementation}

\paragraph{Environment and instrumentation.}
Each session runs in a disposable Ubuntu container with a virtual display, session bus, and accessibility bus. Its seed determines a synthetic filesystem with 12--24 top-level directories, subtrees of varying depth, and valid files of different media types. The policy selects interactable elements from the accessibility tree. Each step records a cursor-free screenshot, absolute cursor coordinates, the target component, the interaction verb, and structured state changes.

\paragraph{State recovery and graph construction.}
The window system identifies the active window and its geometry, while the application identifies the displayed directory. Path recovery uses, in order, the application-reported location, the navigation breadcrumb, and matching between visible labels and filesystem contents. Disagreement lowers confidence and emits a warning without replacing measured values with heuristic estimates. Consecutive post- and pre-states use the same source and settling point. Views are identified by folder path or dialog type; closed transient overlays become placeholder handles, composite icon-label rows become single elements, and decorative widgets are removed. We store the initial graph, subsequent updates, and element bounds for loss weighting and evaluation.

\paragraph{Recovery and validation.}
The collector distinguishes principal windows, transient overlays, popup menus, and foreign windows. When collection stalls, it applies a bounded sequence of dismissal actions, each recorded as a transition. Before each trajectory, it restores the filesystem snapshot and resets windows, history, bookmarks, and geometry. Parallel sessions record their seeds and configurations. Validation checks state continuity, overlay loops, stale target geometry, and missing frames; trajectories with fatal defects are excluded.

\subsubsection{Compared corpora and transition statistics}
\label{app:data-richness}

\paragraph{Compared corpora.}
Section~\ref{exp:data} compares three collection settings:
\begin{itemize}
\item \textbf{Grounded:} 1,956 training trajectories containing 586,210 actions, with screenshots, target identities, semantic outcomes, and graph updates. A separate corpus contains 240 trajectories and 72,000 actions and provides an additional coverage estimate.
\item \textbf{NeuralOS-style:} because the original data was unavailable, we run the published collector unmodified in its native environment \citep{rivard_neuralos_2025}. The resulting 200 clips last 30 seconds each at 15 fps and $512\times384$, yielding 89,800 transitions with frames and raw inputs but no element or state labels. The two cursor-speed modes are 2.5 and 190 pixels/frame, compared with the originally reported 1.5 and 195 at twice the resolution.
\item \textbf{Blind in our environment (B1):} a NeuralOS-style sampler chooses a random key with probability 0.10; otherwise, it clicks uniformly over the $640\times480$ screen, with single/double/right-click probabilities of 65/30/5\%. Inputs are issued through \texttt{xdotool} in a fresh Nautilus container per trajectory; screenshots are captured with \texttt{scrot} after a 0.7-second settling period. We collect 24 trajectories of 300 actions (7,200 actions). Recovery and validity checking are disabled. Records contain screenshots, inputs, and the active-window registry.
\end{itemize}
The last comparison holds the environment fixed, but includes differences in recovery and validation as well as action selection. Metrics requiring unavailable annotations are not reported.

\begin{figure}[htbp]
\centering
\includegraphics[width=0.85\linewidth]{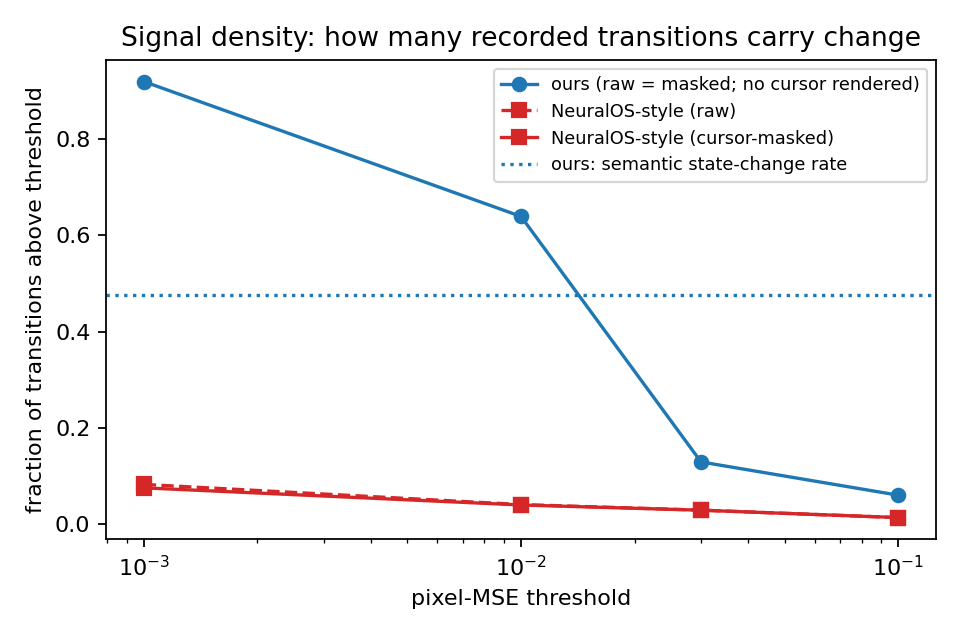}
\caption{\textbf{Visual-change rates across thresholds.} Fraction of transitions meeting each pixel-MSE threshold for grounded collection and NeuralOS-style collection, with and without cursor masking. The dotted line indicates the grounded semantic state-change rate (47.7\%).}
\label{fig:signalDensity}
\end{figure}

\paragraph{Semantic and visual changes.}
The semantic state-change rate is
\begin{equation}
\rho_{\mathrm{sem}}=\frac{1}{n}\sum_{t=1}^{n}
\mathbf{1}[e_t\in\mathcal{C}],
\label{eq:semrate}
\end{equation}
where $e_t$ is the recorded outcome and $\mathcal{C}$ includes navigation, menu opening, and window-geometry changes. Pixel change is the mean squared difference between consecutive frames with intensities scaled to $[0,1]$. We report the proportion $\rho_{\mathrm{pix}}(\tau)$ with MSE at least $\tau$ (Figure~\ref{fig:signalDensity}).

\paragraph{Cursor handling.}
Grounded and B1 screenshots exclude the pointer, so their MSE is computed without masking. This was verified on recorded no-change steps with large cursor displacements, which contain zero changed pixels. NeuralOS-style recordings include the pointer, so we compute both raw and masked MSE. Using the recorded cursor coordinates, we exclude the union of two axis-aligned $32\times32$ boxes centered at the previous and current cursor positions, clipped to the image bounds. We average squared RGB differences over the remaining pixels only. Cursor-only threshold crossings satisfy $\mathrm{MSE}_{\mathrm{raw}}\geq\tau$ and $\mathrm{MSE}_{\mathrm{masked}}<\tau$. In NeuralOS-style data, 79.6\% of consecutive frame pairs have masked MSE below $10^{-5}$; this avoids treating pointer motion as novel interface content.

\paragraph{Transition rates and uncertainty.}

Grounded collection yields 47.7\% semantic state changes and 63.9\% pixel changes at $\tau=10^{-2}$, compared with 4.0\% for cursor-masked NeuralOS-style data and 21.3\% for blind collection in our environment. In NeuralOS-style data, 52.4\% of frames contain only mouse movement; 0.9\% of these meet the masked threshold. Among left-clicks, right-clicks, and keypresses, respectively 9.7\%, 40.2\%, and 3.71\% meet the same threshold. Thus, its low visual-change rate reflects both mouse-only frames and inputs with little visible effect. For signal-density estimates, reported 95\% confidence intervals use 2,000 trajectory-level bootstrap replicates, recomputing the step-count-weighted mean and taking the 2.5th and 97.5th percentiles \citep{efron1993introduction}.

\begin{table}[hbtp]
\centering\small
\caption{\textbf{Collection statistics.} Pixel changes use an MSE threshold of $10^{-2}$; NeuralOS-style rates exclude cursor motion. Dashes indicate unavailable or unreported measurements. Blind-click hit probabilities are geometric estimates over recorded states, not empirical success rates of the blind trajectories. The blind state-count reference ($\dagger$) is a conservative comparison value, not a measured mean.}
\label{tab:datasetRichness}
\begin{tabular}{lccc}
\toprule
& \textbf{Grounded} & \textbf{NeuralOS-style} & \textbf{Blind (our env.)} \\
\midrule
Recorded actions/transitions & 586,210 & 89,800 & 7,200 \\
Semantic state-change rate & 47.7\% & --- & --- \\
Pixel-change rate & 63.9\% & 4.0\% & 21.3\% \\
Interactable-element hit probability & 100\% & --- & 47.8\% \\
State-advancing-element hit probability & --- & --- & 28.7\% \\
Unique states in corpus & 5,615 & --- & --- \\
Unique states per trajectory & 37.1 & --- & $\approx20^{\dagger}$ \\
State-visit entropy (bits) & 6.70 & --- & --- \\
Chao1 estimated support coverage & 88.2\% & --- & --- \\
Reached depth $\geq3$ in 300 actions & 58.6\% & --- & 33.3\% \\
Reached depth $\geq5$ in 300 actions & 18.3\% & --- & 0\% \\
\bottomrule
\end{tabular}
\end{table}

\paragraph{Blind-click geometry.}
For a uniform click, the probability of hitting an interactable element is
\begin{equation}
P_{\mathrm{hit}}(s)=
\frac{\left|\bigcup_{e\in E(s)}R_e\right|}{|\mathcal{S}|},
\label{eq:hitprob}
\end{equation}
where $R_e$ is the element's rectangle clipped to screen $\mathcal{S}$, with overlaps counted once. Averaged over recorded steps, this probability is 47.8\% for any interactable element and 28.7\% for elements observed to advance state. The latter falls from 34\% at the home directory to 7.9\% at depth three and 1.8\% at depth five (Figure~\ref{fig:depthPanels}). Grounded selection targets valid elements by construction, although a valid interaction need not change state.

\subsubsection{Observed exploration and coverage}

\paragraph{State diversity and event reachability.}
The grounded training corpus contains 5,615 distinct states, averaging 37.1 per trajectory, with pooled state-visit entropy of 6.70 bits. The estimated blind state count is 9.3 per trajectory; for comparison, we use a conservative reference of approximately 20 states per 300 actions, favoring the blind baseline. Within 300 actions, grounded collection reaches depths three and five in 58.6\% and 18.3\% of trajectories, compared with 33.3\% and 0\% for blind collection in the same environment.

For each event $G$, we measure its first occurrence $\tau_G=\min\{t:G\text{ occurs at }t\}$, right-censored at 300 actions. Reachability and median occurrence time among successful trajectories are: context menus, 98.5\% (step 14); drags, 98.4\% (11); typing or renaming, 96.9\% (30); overlays or dialogs, 97.7\% (20); and filesystem mutations, 86\% (7).

\begin{figure}[hbtp]
\centering
\includegraphics[width=\linewidth]{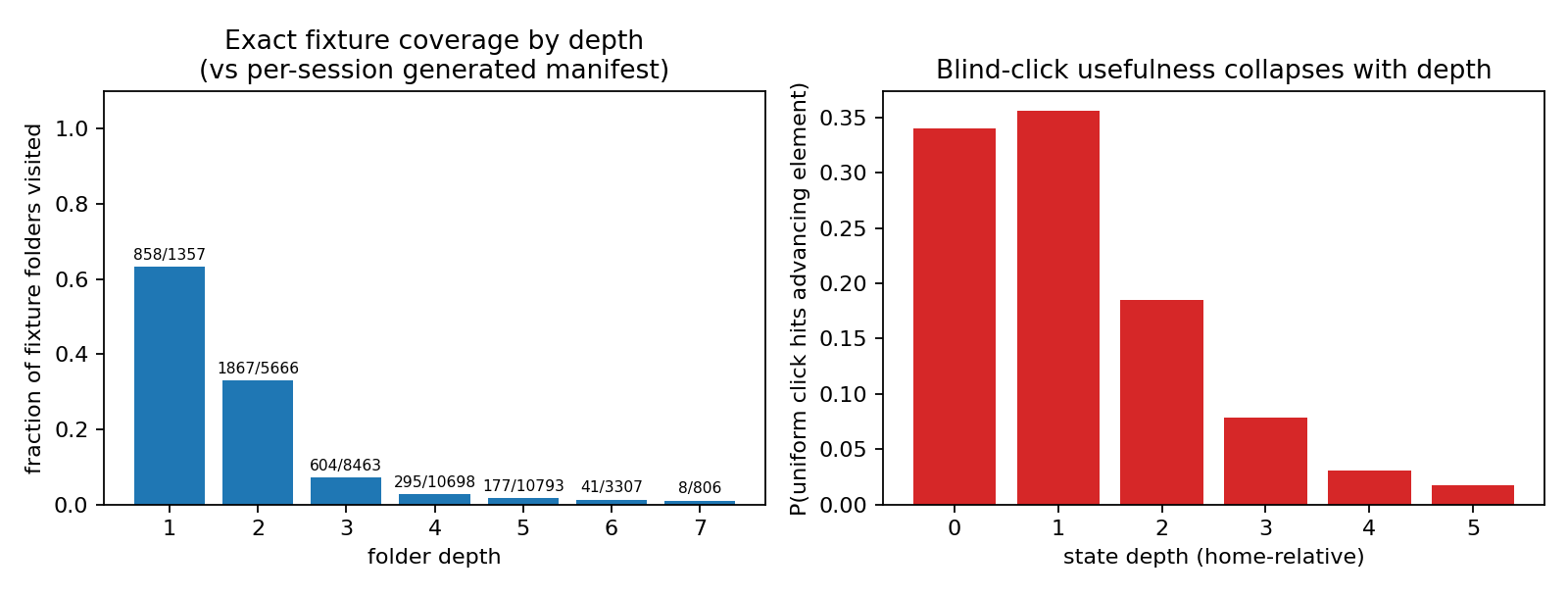}
\caption{\textbf{Coverage and click geometry by depth.} Left: folder coverage relative to each session's generated filesystem. Right: probability that a uniform click hits an element observed to advance state.}
\label{fig:depthPanels}
\end{figure}

\paragraph{Filesystem coverage and sampled state support.}
Across 60 sessions, each comprising twenty 300-action trajectories against its own seeded filesystem, mean folder coverage is 9.6\% with a standard deviation of 2.8 percentage points across sessions (computed with divisor 60). This spread describes session-to-session variation, not uncertainty in the mean. Coverage ranges from 63\% at depth one to approximately 1\% at depth seven (Figure~\ref{fig:depthPanels}). It measures exploration of the complete generated filesystem, separately from the sampled-support estimate below.

\begin{figure}[hbtp]
\centering
\includegraphics[width=0.8\linewidth]{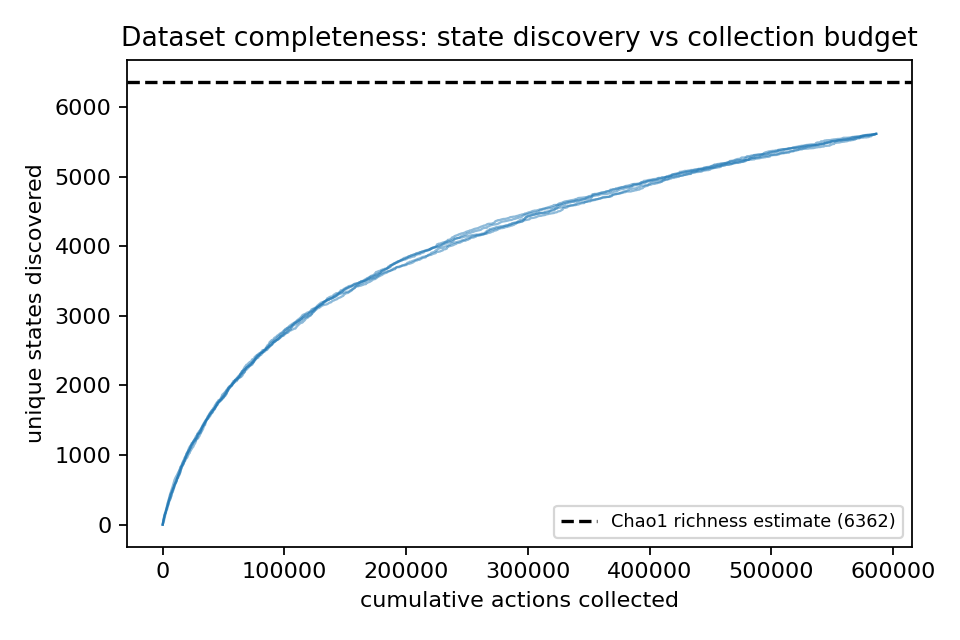}
\caption{\textbf{Observed state accumulation.} Unique states against cumulative actions over five shuffled trajectory orders. The dashed line indicates the Chao1 support estimate, $\widehat{S}\approx6362$.}
\label{fig:rarefaction}
\end{figure}

We estimate sampled state support using Chao1 \citep{chao1984nonparametric},
\begin{equation}
\widehat{S}=S_{\mathrm{obs}}+\frac{f_1^2}{2f_2},
\label{eq:chao1}
\end{equation}
where $S_{\mathrm{obs}}$ is the number of distinct observed states and $f_1,f_2$ count states visited once and twice. Training yields $S_{\mathrm{obs}}=5615$, $f_1=890$, $f_2=530$, and $\widehat{S}\approx6362$, corresponding to 88.2\% estimated support coverage; validation yields 89.1\%. Dependence between visits within trajectories limits interpretation of these estimates. They do not establish coverage of all reachable environment states. Figure~\ref{fig:rarefaction} shows state accumulation over five shuffled trajectory orders.

\subsubsection{Policy comparison on an estimated transition model}

\paragraph{Model and policies.}
To examine the effect of action selection, we simulate navigation on a transition model estimated from the grounded corpus. States are keyed by folder path, and content mutations are approximated as preserving identity. Each state supplies observed elements $E(s)$, their screen areas, and destinations $\widehat{T}(s,e)$, resolving disagreements by majority vote. Unobserved transitions are treated as self-loops under every policy. This approximation omits dialog traps and limits exploration to transitions represented in the corpus.

We compare four policies: blind sampling with the NeuralOS-style input mix, blind sampling with one click per step, and uniform or production-weighted grounded selection. The first samples a per-walk click probability uniformly on $[0,0.4]$; non-click steps self-loop. Both blind policies sample screen positions uniformly and use a double click with probability 0.3 when the destination requires one. Grounded policies select elements and use the required click type. Each policy runs 2,000 walks of 300 steps from the home directory (Algorithm~\ref{alg:walk}).

\begin{algorithm}[hbtp]
\caption{One step on the estimated transition model}
\label{alg:walk}
\begin{algorithmic}[1]
\Require state $s$, elements $E(s)$, destinations $\widehat{T}$, policy $\pi$, blind click probability $c$
\If{$\pi$ is blind}
  \State with probability $1-c$: \Return $s$
  \State sample a screen position uniformly; identify the hit element $e$
  \If{no element is hit}
    \State \Return $s$
  \EndIf
  \If{$e$ requires a double click}
    \State with probability $0.7$: \Return $s$
  \EndIf
\Else
  \State sample $e\in E(s)$ uniformly or using production weights
\EndIf
\If{$\widehat{T}(s,e)$ is undefined}
  \State \Return $s$
\EndIf
\State \Return $\widehat{T}(s,e)$
\end{algorithmic}
\end{algorithm}

\begin{figure}[htbp]
\centering
\includegraphics[width=0.9\linewidth]{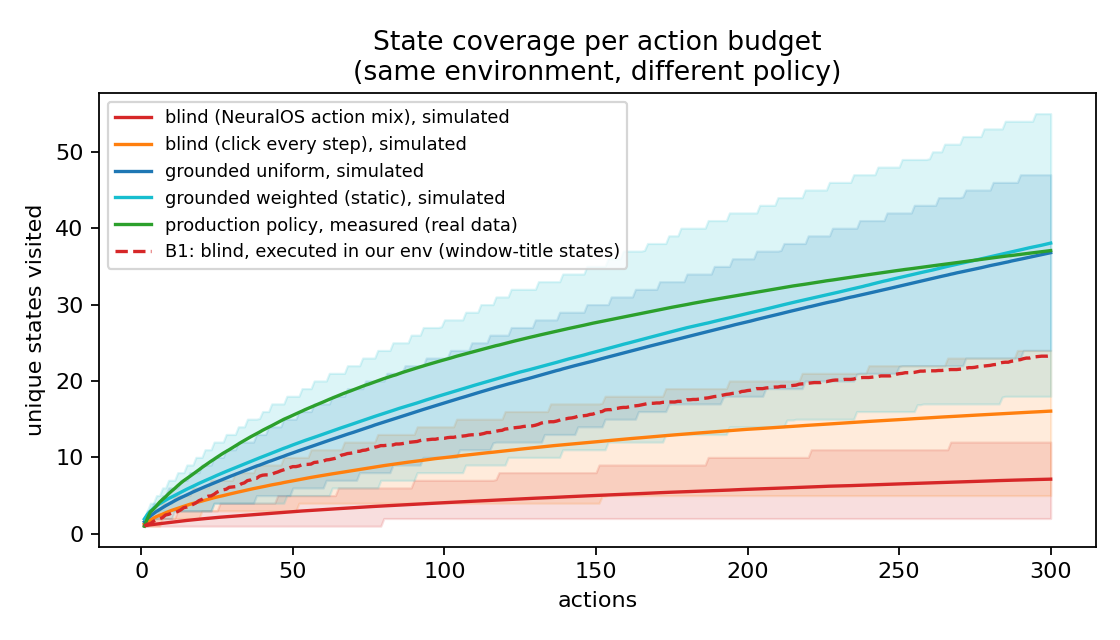}
\caption{\textbf{Exploration under simulated policies.} Unique states against action budget, with means and 10th--90th percentile bands over 2,000 walks. The measured grounded count and conservative blind reference are shown for comparison. Simulations use corpus-estimated navigation transitions and omit dialog traps.}
\label{fig:coveragePolicies}
\end{figure}

\paragraph{Simulation results.}
The four policies visit respectively 7.1, 16.1, 36.8, and 38.1 unique states per 300 steps (Figure~\ref{fig:coveragePolicies}). The blind input mixture reaches depth three in 88\% of simulated walks, with median first arrival at step 46, compared with every walk and median 1--2 steps for both grounded policies. Depth-five reachability is 0.8\% for the blind mixture and 78.9\% for weighted grounded selection. Pooled visit entropies are 5.67, 8.75, and 10.62 bits for the blind mixture, uniform grounded, and weighted grounded policies, respectively.

These estimates describe walks on the corpus-derived navigation model, not measured environment executions. Actual collection reaches 37.1 states per trajectory under the grounded policy, but its depth-five reachability is 18.3\%, illustrating the limits of the simplified model. Transient views and their state-counting conventions must also be distinguished when comparing simulations with real trajectories.

\subsection{Experiments}
\label{app:extra}

\subsubsection{Memory Experiment}
\label{app:mem}
In the memory experiment of Section~\ref{exp:mem}, the target's first visit occupies frames 7-9, with zero-based indexing. Frame 7 belongs to the ground-truth context, whereas frames 8-9 are generated. For self-consistency, we pair each of these two generated frames with each of the five target-revisit frames, producing 10 pairs per trajectory, 1,840 pairs per horizon, and 9,200 pairs overall. Both models are evaluated at $256\times192$ resolution using the same element-crop matching and SSIM procedures, yielding 12,510 crop comparisons per horizon with no skipped comparisons. Only the designated target is scored: revisits to ancestor folders encountered during distraction are excluded. This benchmark evaluates retention of unchanged content; it does not test memory updates following content mutations.

\setcounter{secnumdepth}{4}

\paragraph{Reconstruction correctness}
\label{app:revis}

\begin{figure}[t]
    \centering
    \includegraphics[width=0.65\linewidth]{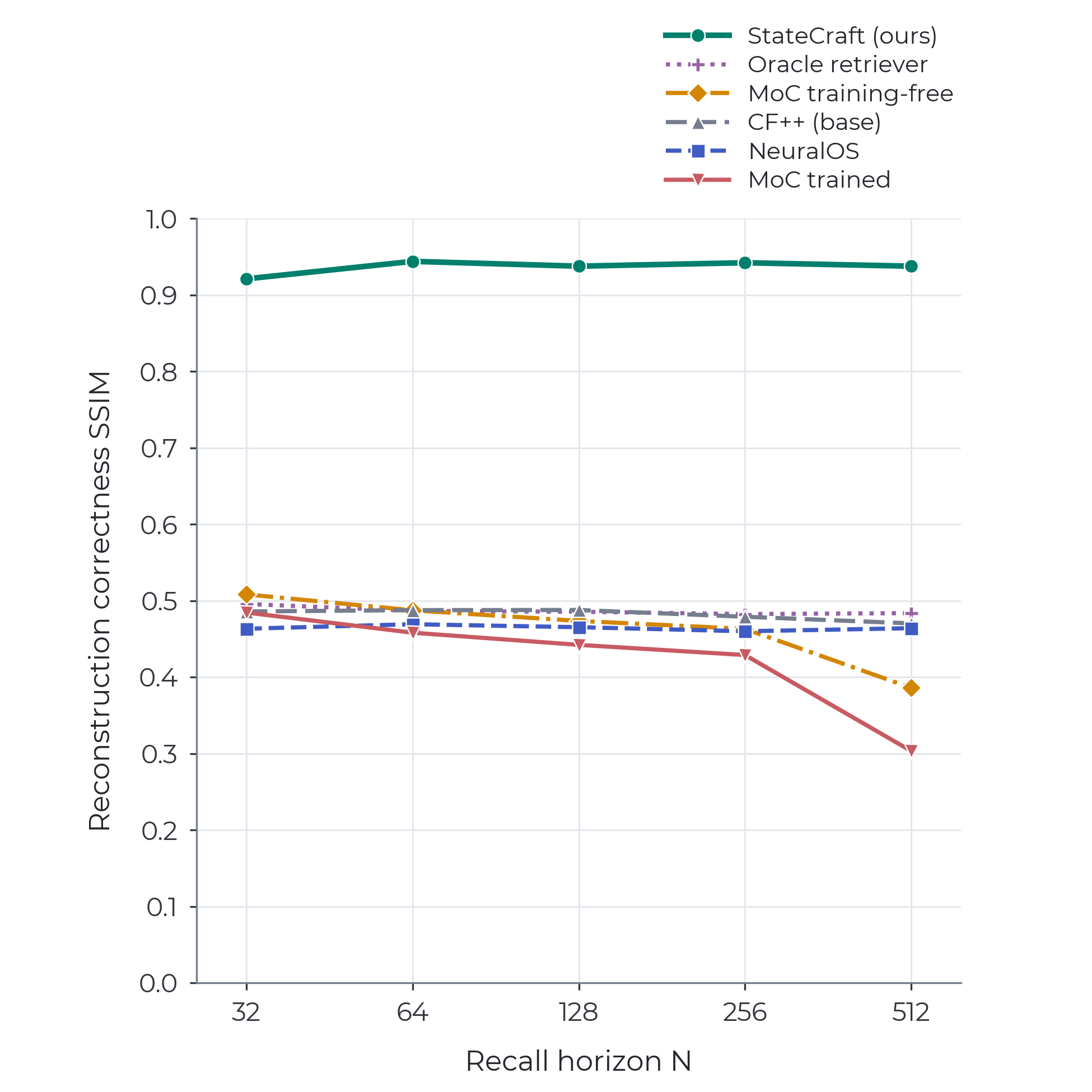}
    \caption{\textbf{Revisit reconstruction correctness.}
    ROI-level SSIM against ground truth for the designated target.
    \ours{} maintains high reconstruction accuracy across horizons,
    complementing the self-consistency results in Figure~\ref{fig:memory}.}
    \label{fig:memory_gt}
\end{figure}

We compare generated target-revisit frames with their corresponding ground-truth frames using ROI-level SSIM. Figure~\ref{fig:memory_gt} shows that \ours{} maintains SSIM of 0.92--0.94 across all horizons, whereas no baseline exceeds 0.51. Oracle retrieval remains close to base CF++ despite improving self-consistency, indicating that selecting the correct past latent under this cache mechanism does not by itself yield correct reconstruction. These results confirm that our high revisit consistency accompanies similarity to the target content.

\paragraph{Mixture of Contexts training details}
\label{app:moc}

We adapt Mixture of Contexts \citep{cai2025mixturecontextslongvideo} to the causal KV-cache implementation of our CF++ baseline, without graph conditioning. Historical keys are partitioned into one-frame chunks of 1,560 tokens. A parameter-free router scores chunks using dot products between pooled query and key representations. To accommodate the frame-autoregressive implementation, routing decisions are shared across the query tokens within each frame, rather than computed separately for each token. We set the chunk-selection parameter to 21 and retain the two most recent chunks as mandatory context. Text remains accessible through Wan's cross-attention. During training, selected non-mandatory chunks undergo context drop-off and drop-in, using $p_{\max}=0.5$ and $\lambda=2$, respectively (these values are implementation choices).

Starting from the CF++ checkpoint trained without graph conditioning, we finetune for 4,500 optimizer steps on eight H200 GPUs, with a global batch size of eight and no gradient accumulation. Training uses 76-latent-frame clips at $640\times480$ resolution and the base model's consistency-distillation objective, with learning rate $2\times10^{-6}$, weight decay 0.01, and EMA decay 0.99.

We evaluate checkpoints at steps 1,250 and 4,500, together with a training-free variant that applies the same routing configuration to the original CF++ checkpoint. In Figure~\ref{fig:memRecall} we plot the best finetuned checkpoint (step 1250), and the training-free adaptation. This comparison separates the effect of selecting historical context from that of finetuning the backbone for retrieval. Without finetuning, routing improves the self-consistency of the revisits over the rolling-window baseline at horizons $N=32$--$256$, but scores lower at $N=512$. Finetuning does not consistently improve these scores and progressively reduces visual sharpness. At $N=512$, self-consistency SSIM is 0.5304 without finetuning, 0.4127 at step 1,250, and 0.3588 at step 4,500.

\paragraph{Oracle retrieval.}
\label{app:oracle}
We augment CF++ with an episode-specific bank indexed by ground-truth folder identity, storing the first generated latent for each folder and excluding ground-truth context frames. On subsequent occurrences, the stored latent replaces the newly generated latent only when updating the context cache; output frames remain model-generated. Retrieval therefore influences subsequent frames, not the first frame of a revisit. Retrieved latents use the current temporal position, and all other generation settings match base CF++. This baseline removes state-matching errors through privileged identities unavailable at deployment, while retaining first-visit generation errors. It tests oracle selection under this cache mechanism, rather than an upper bound on retrieval-based generation generally.

\setcounter{secnumdepth}{3}

\subsubsection{ROI scores for DFoT}
\label{app:dfot}

In Table~\ref{tab:elemMetricsDFoT} we expand the evaluation we present in Section~\ref{exp:consist} using DFoT as our backbone.

\begin{table}[hbtp]
\centering
\caption{\textbf{Render quality by screen region for the DFoT backbone, over the full region (top) and inside OCR-detected text boxes only (bottom).} Gains over the baselines are largest on the canvas, where filesystem state is displayed, and largest of all on canvas text, which the baselines fail to render legibly. On the static side panel, all models perform comparably. Last row of each block: upper bound with ground-truth diffs.}
\label{tab:elemMetricsDFoT}
\begin{tabular}{lcccccc}
\toprule
& \multicolumn{2}{c}{\textbf{Breadcrumb}} & \multicolumn{2}{c}{\textbf{Canvas}} & \multicolumn{2}{c}{\textbf{Side panel}} \\
\cmidrule(lr){2-3} \cmidrule(lr){4-5} \cmidrule(lr){6-7}
\textbf{Model} & SSIM & PSNR & SSIM & PSNR & SSIM & PSNR \\
\midrule
\multicolumn{7}{l}{\footnotesize\itshape Full region} \\
NeuralOS & \underline{0.85} & 30.12 & 0.61 & 19.74 & \textbf{0.96} & \underline{34.28} \\
DFoT & 0.83 & \underline{31.27} & \underline{0.62} & \underline{21.48} & \underline{0.95} & \textbf{34.71} \\
\ours{} (DFoT) & \textbf{0.91} & \textbf{32.88} & \textbf{0.89} & \textbf{31.83} & 0.92 & 33.81 \\
\textit{\ours{} (DFoT), GT diffs} & 0.93 & 33.41 & 0.92 & 32.87 & 0.92 & 33.92 \\
\midrule
\multicolumn{7}{l}{\footnotesize\itshape OCR-detected text boxes} \\
NeuralOS & \underline{0.52} & 27.16 & 0.15 & 16.71 & \underline{0.80} & 27.90 \\
DFoT & 0.39 & \underline{29.51} & \underline{0.20} & \underline{19.10} & \textbf{0.85} & \textbf{30.47}\\
\ours{} (DFoT) & \textbf{0.74} & \textbf{31.20} & \textbf{0.81} & \textbf{26.55} & 0.79 & \underline{29.55} \\
\textit{\ours{} (DFoT), GT diffs} & 0.79 & 31.60 & 0.91 & 27.60 & 0.88 & 29.60 \\
\bottomrule
\end{tabular}
\end{table}

\subsubsection{State-predictor evaluation}
\label{app:llm}

The state-predictor module is scored here against the ground truth recorded on node
accuracy (exact structural match of the next node), transition accuracy, schema validity, and generation speed in tokens per second. Table~\ref{tab:llmEval} shows the evaluation of two backbones for in-domain and out-of-domain tasks. The OOD tasks are either unseen scenarios following the same JSON structure (i.e. memory game, battleship, blackjack) and unseen operations (i.e. "delete all the files starting with the letter 'A'"). For this evaluation, we test the model on the graph states of the hold-out set for the consistency experiment (Section~\ref{exp:consist}), and on a set of synthetic experiments for the OOD scenarios.

\begin{table}[h]
\centering
\small
\setlength{\tabcolsep}{4pt}
\caption{\textbf{State predictor evaluation: finetuned (FT) against base with few-shot prompts (FS), both backbones on the same corpus and schedule.} In-domain: node accuracy, transition accuracy and schema validity on held-out navigation. OOD: node accuracy on unseen operations and cross-domain games. Finetuning Qwen3.5-0.8B trades generality for in-domain accuracy; DiffusionGemma-26B scores are lower over the task-specific metrics, but it maintains generalization capabilities and decodes faster.}
\label{tab:llmEval}
\begin{tabular}{llcccccc}
\toprule
& & \multicolumn{3}{c}{\textbf{In-domain}} & \multicolumn{2}{c}{\textbf{OOD}} & \\
\cmidrule(lr){3-5} \cmidrule(lr){6-7}
\textbf{Model} & & Node & Trans. & Schema & Ops & Games & \textbf{TPS} \\
\midrule
Qwen3.5-0.8B~\citep{qwen35blog} & FT & \textbf{0.86} & \textbf{0.96} & \textbf{0.91} & 0.00 & 0.18 & 250 \\
 & FS & 0.48 & 0.52 & 0.85 & \textbf{1.00} & 0.83 & 250 \\
DiffusionGemma-26B~\citep{diffusiongemmateam2026diffusiongemmatechnicalreport} & FT & 0.77 & 0.85 & 0.79 & \textbf{1.00} & \textbf{0.86} & \textbf{1500} \\
 & FS & 0.34 & 0.40 & 0.70 & \textbf{1.00} & 0.85 & \textbf{1500} \\
\bottomrule
\end{tabular}
\end{table}

\subsubsection{Full ablation results}
\label{app:cond-ablation}
 
\paragraph{Conditioning representation.}
We evaluate two reduced conditioning variants of DFoT while keeping the backbone, frozen encoder, cross-attention mechanism, and training budget fixed. \textit{Path} embeds only the current filesystem path into one token per frame, providing state identity without contents. \textit{Flat-node} embeds the entire serialized node, including its path and visible elements, into one token, retaining content without separate element representations. Both single-token variants retain the encoder's native embedding width, whereas the full model truncates embeddings to accommodate multiple graph tokens. Both variants are finetuned from the same action-conditioned checkpoint using the full model's warm-up and training schedule. Together with the unconditioned baseline and full model, they test the contributions of state identity, content, and elementwise access.

Table~\ref{tab:degrad} shows that path conditioning improves breadcrumb reconstruction but provides little benefit for canvas content or revisit consistency. Flat-node conditioning improves both, while separate graph tokens yield the strongest results: canvas SSIM increases from 0.62 without conditioning to 0.76 with flat-node conditioning and 0.92 with graph tokens. Within canvas text boxes, the corresponding scores are 0.20, 0.54, and 0.91. Graph tokens also improve canvas PSNR by $\sim$7 dB over flat-node conditioning and maintain revisit scores of 0.89--0.93 across the evaluated distances. The unconditioned model retains a small advantage on full-region side-panel metrics, where appearance is identical across trajectories.

\begin{table}[t]
\centering
\small
\setlength{\tabcolsep}{4pt}
\renewcommand{\arraystretch}{0.95}
\caption{\textbf{Conditioning representation ablation on DFoT.}
Reconstruction cells report SSIM / PSNR (dB); text scores use
OCR-detected boxes. Memory scores measure revisit similarity by
action lag, with approximately 0.5 indicating plausible but unrelated
content. None denotes no state conditioning. All models receive actions, and all conditioning variants use ground-truth states, isolating the representation supplied to the video model.
Best values are bold and second-best underlined, including ties.}
\label{tab:degrad}
\label{tab:degrad1}
\label{tab:degrad2}
\label{tab:degrad3}
\begin{tabular}{@{}lcccc@{}}
\toprule
\textbf{Region / lag}
& \textbf{None}
& \textbf{Path}
& \textbf{Flat-node}
& \textbf{Graph (full)} \\
\midrule
\multicolumn{5}{l}{\textit{Full-region reconstruction: SSIM / PSNR}} \\
Breadcrumb
& 0.83 / 31.27
& \underline{0.92} / 31.18
& 0.91 / \underline{32.55}
& \textbf{0.93} / \textbf{33.41} \\
Canvas
& 0.62 / 21.48
& 0.62 / 21.80
& \underline{0.76} / \underline{25.92}
& \textbf{0.92} / \textbf{32.87} \\
Side panel
& \textbf{0.95} / \textbf{34.71}
& \underline{0.94} / \underline{34.29}
& 0.93 / 33.70
& 0.92 / 33.92 \\
\midrule
\multicolumn{5}{l}{\textit{Text reconstruction: SSIM / PSNR}} \\
Breadcrumb
& 0.39 / 29.51
& \textbf{0.80} / \textbf{31.65}
& 0.41 / 30.95
& \underline{0.79} / \underline{31.60} \\
Canvas
& 0.20 / 19.10
& 0.20 / 19.60
& \underline{0.54} / \underline{22.42}
& \textbf{0.91} / \textbf{27.60} \\
Side panel
& \underline{0.85} / \underline{30.47}
& 0.82 / \textbf{30.49}
& 0.82 / 29.69
& \textbf{0.88} / 29.60 \\
\midrule
\multicolumn{5}{l}{\textit{Revisit memory by action lag: SSIM}} \\
8--16  & 0.52 & 0.53 & \underline{0.68} & \textbf{0.89} \\
16--24 & 0.56 & 0.56 & \underline{0.69} & \textbf{0.92} \\
24--32 & 0.54 & 0.54 & \underline{0.72} & \textbf{0.93} \\
32--64 & 0.58 & 0.58 & \underline{0.70} & \textbf{0.92} \\
64+    & 0.55 & 0.56 & \underline{0.70} & \textbf{0.93} \\
\bottomrule
\end{tabular}
\end{table}
 
\paragraph{Conditioning ablation.}
We test whether the gains in Section~\ref{exp:consist} depend on graph conditioning beyond the ROI-weighted loss, and whether actions remain useful when the target state is supplied explicitly. For both DFoT and CF++, we independently zero the graph or action channel at inference, following the independent conditioning dropout used during training. Evaluating the same finetuned models holds the learned weights and training objective fixed while measuring their dependence on each signal (Table~\ref{tab:ablations}).

Removing graph conditioning substantially degrades content reconstruction and memory: canvas SSIM falls by 0.333 for CF++ and 0.291 for DFoT, while revisit SSIM decreases by 0.222 and 0.210, respectively. These losses occur despite training with ROI reweighting, showing that the reweighted objective alone does not sustain the full model's state fidelity when graph information is unavailable. This experiment establishes the model's dependence on graph conditioning, although it does not separately quantify the contribution of loss reweighting.

Removing actions produces smaller reconstruction losses across both backbones, but its effect is not negligible: canvas SSIM decreases by 0.061 for CF++ and 0.051 for DFoT. Revisit consistency remains nearly unchanged for CF++ but drops by 0.092 for DFoT. Qualitative rollouts retain approximately correct target contents while exhibiting noisier, less plausible transitions, consistent with actions helping the model apply visual changes. Together, these results support complementary roles for the two channels: graph conditioning supplies explicit state contents, while actions improve their realization during interaction.

\begin{table}[t]
\centering\small
\setlength{\tabcolsep}{4pt}
\caption{\textbf{Effect of the different conditioning signals on the main metrics.} Each channel is removed at inference
by zeroing it, on the same 240 test trajectories with an 8-frame
ground-truth seed and closed-loop generation. Graph tokens carry \emph{what} is
on screen: removing them costs the canvas 0.33 (CF++) and 0.29 (DFoT) SSIM and
roughly a fifth of revisit memory, and drops breadcrumb agreement with the
ground-truth path by 14--19 points. Removing the action channel is far cheaper
on both backbones. Compare columns within a backbone: since the ablation is meant to compare the effectiveness of the components within an architecture, both CF++ and DFoT are scored here at their
native resolution ($640{\times}480$ and$256{\times}192$).}
\label{tab:ablations}
\begin{tabular}{l ccc ccc}
\toprule
& \multicolumn{3}{c}{CF++} & \multicolumn{3}{c}{DFoT} \\
\cmidrule(lr){2-4}\cmidrule(lr){5-7}
SSIM $\uparrow$ & full & $-$act. & $-$graph & full & $-$act. & $-$graph \\
\midrule
Whole frame       & 0.934 & 0.895 & 0.824 & 0.929 & 0.889 & 0.857 \\
Canvas            & 0.897 & 0.836 & 0.564 & 0.907 & 0.856 & 0.616 \\
Breadcrumb        & 0.890 & 0.855 & 0.759 & 0.941 & 0.922 & 0.833 \\
Semantic (all)    & 0.891 & 0.859 & 0.788 & 0.946 & 0.916 & 0.876 \\
Revisit $k{=}64$  & 0.878 & 0.874 & 0.656 & 0.898 & 0.806 & 0.688 \\
\midrule
Breadcrumb match rate $\uparrow$ & 0.905 & 0.872 & 0.769 & 0.932 & 0.904 & 0.812 \\
\bottomrule
\end{tabular}
\end{table}

\subsubsection{Latent space probing}
\label{app:probe-setup}

We test whether video-model activations predict the next UI transition before language-model invocation. With CF++ frozen, we extract residual activations from Wan block~15 of 30 during teacher-forced evaluation, at diffusion timestep $\tau{=}1000$, the first step of the two-step sampler. Global, cursor-token, and cursor-neighborhood pooling of the $30\times52$ latent grid produce three 1536-dimensional vectors, concatenated with eight action features into a 4616-dimensional representation.

Controls include actions alone; symbolic features describing the clicked graph slot, its kind, name, area, sidebar membership, and valid-slot count; pixels from a decoded-frame thumbnail and cursor crop; and pooled and cursor-aligned graph-conditioning features. Including actions, their dimensions are 8, 26, 200, and 2056, respectively. Each representation is evaluated with multinomial logistic regression and a two-layer GELU MLP with dropout, predicting \textsc{no\_change}, \textsc{up\_back}, \textsc{window\_close}, \textsc{deeper}, or \textsc{jump}. Trajectories are partitioned into disjoint training, calibration, and test sets. Calibration determines early stopping and gate thresholds; Table~\ref{tab:cf-probe} reports results on 35{,}880 held-out test decisions.

Denoiser features achieve 0.987 macro AUROC and 0.916 five-class accuracy, versus 0.496 majority-class accuracy. Linear accuracy reaches 0.901. Latents outperform graph conditioning (0.870 accuracy), symbolic features (0.835), and actions alone (0.790), but only slightly exceed pixels (0.911 accuracy; 0.985 AUROC). Thus, transition information is accessible in denoiser activations, although much is also recoverable from decoded frames.

For deployment, the gate combines activations cached while generating $x_t$ with incoming action $a_t$. Thresholding its predicted \textsc{no\_change} probability determines whether to invoke the language model; the updated or retained state then conditions $x_{t+1}$. This uses CF++ block~15 activations, or DFoT's mid07 activations, without an additional backbone pass. Only the cached-latent configuration is deployed.

At a threshold calibrated for 99\% skip purity, the held-out probe skips 41.0\% of decisions with 99.0\% purity, defined as the fraction of skips requiring no state change. In closed-loop evaluation without a verifier, it skips 3,914 of 9,600 decisions across 48 trajectories (40.8\%), including 345 incorrect skips, giving 91.2\% purity and 5,686 language-model calls. Incorrect skips retain stale states that can affect later decisions. Relative to generation without gating, PSNR decreases by 0.57~dB and SSIM by 0.008. These results establish call savings, not an end-to-end latency improvement.

\begin{table}[t]
\centering\small
\setlength{\tabcolsep}{4.5pt}
\caption{\textbf{Transition probing across feature families.}
Results on 35{,}880 test decisions from trajectories disjoint from training and calibration. All representations include eight action features. We report MLP results and logistic-regression accuracy ($\mathrm{Acc}_5^{\mathrm{lin}}$). AUROC is five-class one-vs-rest macro AUROC; AUROC$_2$ measures change versus no change. $\mathrm{Acc}_5$ is five-class accuracy and F1 is macro-F1. Skip is the fraction of omitted calls; purity is the fraction of skips requiring no state change. Thresholds calibrated for 99\% purity are applied unchanged to test. These are held-out probe results, closed-loop results are reported in the text.
}
\label{tab:cf-probe}
\begin{tabular}{@{}l r cc cc c cc@{}}
\toprule
Features & Dim. & AUROC & AUROC$_2$
& $\mathrm{Acc}_5$ & F1 & $\mathrm{Acc}_5^{\mathrm{lin}}$
& Skip & Purity \\
\midrule
Majority class
& -- & 0.500 & 0.500 & 0.496 & 0.133 & 0.496 & -- & -- \\
\midrule
Action
& 8 & 0.944 & 0.919 & 0.790 & 0.736 & 0.790 & 0.329 & 0.981 \\
Symbolic features
& 26 & 0.965 & 0.941 & 0.835 & 0.805 & 0.801 & 0.395 & \textbf{0.991} \\
Graph conditioning
& 2056 & 0.978 & 0.960 & 0.870 & 0.820 & 0.819 & 0.397 & 0.990 \\
Raw pixels
& 200 & 0.985 & 0.974 & 0.911 & 0.882 & 0.869 & \textbf{0.427} & \textbf{0.991} \\
Denoiser latents
& 4616 & \textbf{0.987} & \textbf{0.975}
& \textbf{0.916} & \textbf{0.898} & \textbf{0.901}
& 0.410 & 0.990 \\
\bottomrule
\end{tabular}
\end{table}

\subsubsection{3D environment experiment details}
\label{app:3dexp}

\paragraph{Data and conditioning.}
We collect 8,644 training, 294 validation, and 828 test trajectories in OmniGibson interiors, with 14 houses represented in training and two unseen houses reserved for testing. The test set contains seven scene groups, each defined by a house and route setup. Trajectories vary camera paths, switch activations, and target-room visits. Each clip contains 81 video frames, encoded into 21 latent frames. The video model receives actions and simulator-derived room-category and illumination-state tokens through the abstract-state conditioning pathway. Room-identity tokens are omitted. State updates are supplied by the simulator in this experiment. Conditioning is aligned with the next generated frame and provided with an eight-frame lead, making the illumination signal available during the ground-truth context before the target room is revealed.

\paragraph{Training and generation.}
We adapt the CF++ backbone using conditioning warmup followed by joint training. The training objective upweights five reveal frames by a factor of 50 and includes a contrastive term that encourages sensitivity to the state signal. We select the checkpoint using the validation split and report results on the held-out test set; the selected checkpoint corresponds to 3,000 joint-training steps. The comparison model starts from the same warmup checkpoint and trains for 3,000 steps with the graph-conditioning pathway absent. Evaluation supplies 14 ground-truth latent frames and autoregressively generates the remaining seven.

\paragraph{Illumination correctness.}
We evaluate 120 test episodes, balanced between lit and dark outcomes and spanning all seven scene groups. Generated frames are decoded to RGB in $[0,1]$ and converted to Rec.\,709 luminance,
$L_t=0.2126R_t+0.7152G_t+0.0722B_t$.
We average spatially and over the video frames associated with each latent frame to obtain $\ell_k$. The episode score is

\[
s=\frac{1}{|\mathcal{W}|}\sum_{k\in\mathcal{W}}\ell_k,
\qquad
\mathcal{W}=\{e,\ldots,T-1\},
\]

where $e$ denotes first entry into the target room, determined from recorded metadata. All 120 evaluated episodes enter at latent frame 15. We compute AUC from these scores and the illumination labels. State-conditioned generation achieves 0.8644 AUC, versus 0.5422 without state conditioning. Accuracy is 0.742 versus 0.542, using a shared threshold equal to the midpoint of the ground-truth lit and dark means, fitted while excluding the evaluated scene group.

\paragraph{Paired state intervention.}
For 240 test episodes, we generate paired continuations with identical visual context and noise, forcing the illumination token to either lit or dark. The mean luminance difference is 0.09218, corresponding to 63.7\% of the ground-truth reference difference of 0.14471, estimated from camera-matched opposite-outcome pairs. The intervention changes brightness in the expected direction in 234 of 240 episodes and all seven scene groups. A two-sided one-sample $t$-test on the seven group-mean differences gives $t(6)=6.58$, $p=0.0006$. Episode counts are descriptive; the significance test uses scene groups as its units.

\subsubsection{Additional Visualizations}
\label{app:vis}

\begin{figure}[htbp]
\centering
\includegraphics[width=\linewidth]{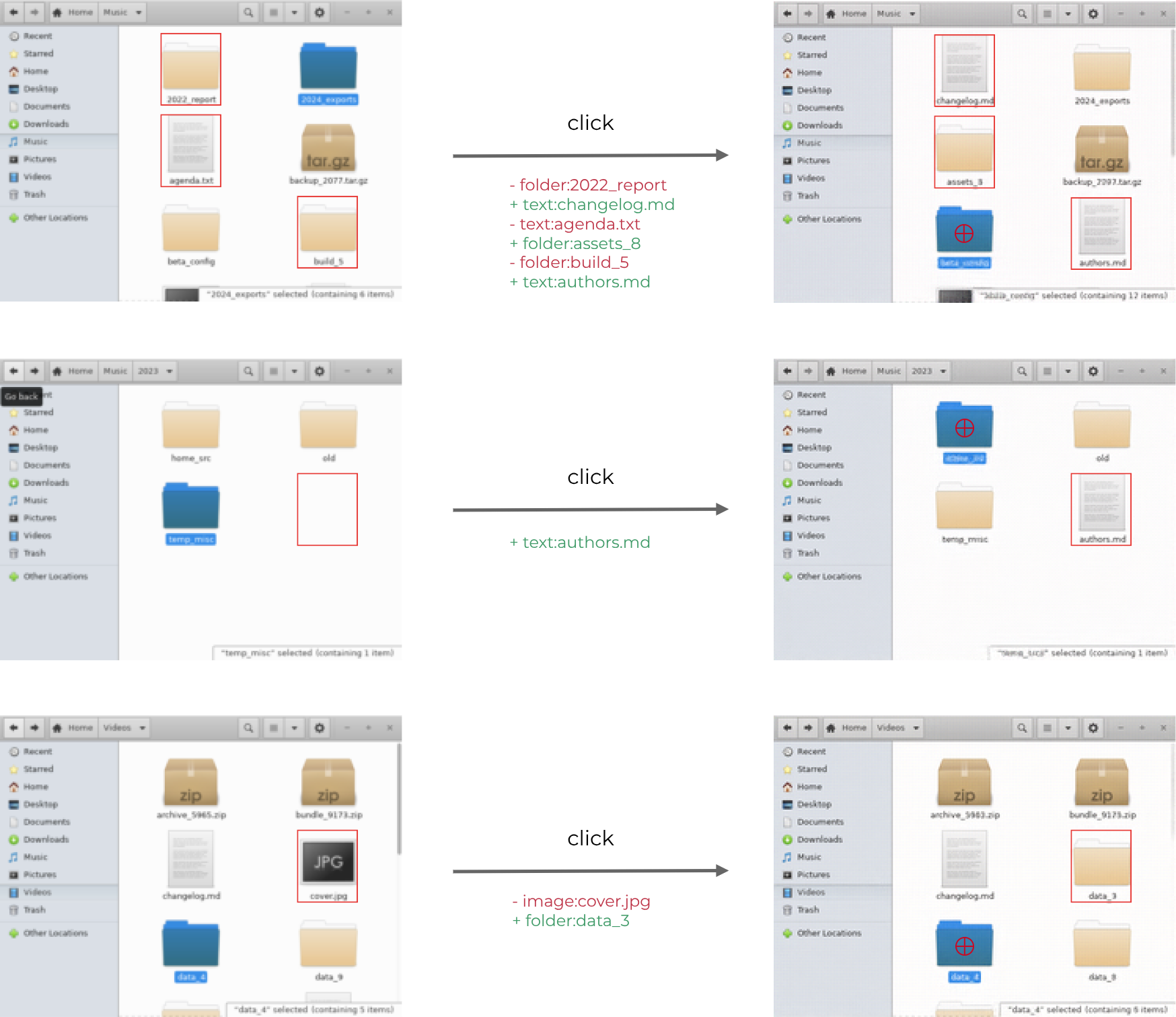}
\caption{\textbf{Controllability interventions by type (multiple edits, new element on background, rename with type change).} Per row: original state (left),
commanded graph diff, removals in red and additions in green (centre), and
the causally generated frame for the edited state, which never occurs in
the data (right; red boxes mark edited regions). The render follows the state
edit rather than the pixels of its own context.}
\label{fig:ctrl}
\end{figure}

\end{document}